%% file: paper.tex
\documentclass{article}
\PassOptionsToPackage{colorlinks=true,allcolors=blue}{hyperref}

\usepackage[english]{babel}
\usepackage[letterpaper,top=2cm,bottom=2cm,left=3cm,right=3cm,marginparwidth=1.75cm]{geometry}

\usepackage{amsmath}
\usepackage{amsfonts}
\usepackage{amssymb}
\usepackage{graphicx}
\usepackage{booktabs}
\usepackage{natbib}
\usepackage{doi}
\usepackage{pdflscape}
\usepackage{microtype}
\usepackage{booktabs}
\usepackage{lscape}
\usepackage{graphicx}
\usepackage{placeins}

\usepackage{longtable}
\usepackage{subcaption}

\usepackage{authblk}
\usepackage{cleveref}
\usepackage[normalem]{ulem}
\usepackage{dsfont}

\newenvironment{leaderboardtable}[2]{%
  \def\leaderboardcaptiontext{#1}%
  \def\leaderboardlabeltext{#2}%
  \begin{table}[htbp]%
  \centering
  \footnotesize
  \setlength{\tabcolsep}{3pt}%
}{%
  \normalsize
  \caption{\leaderboardcaptiontext}%
  \label{\leaderboardlabeltext}%
  \end{table}%
}

\title{ScoringBench: A Benchmark for Evaluating Tabular Foundation Models with Proper Scoring Rules}
\author[1]{Jonas Landsgesell \thanks{jonaslandsgesell\_at\_gmail dot com}}
\author[1]{Pascal Knoll \thanks{knollpascal00\_at\_gmail dot com}}
\author[2,3]{Tizian Wenzel \thanks{wenzel\_at\_math dot lmu dot de}}
\affil[1]{University of Stuttgart (Stuttgart, Germany)}
\affil[2]{Ludwig Maximilian University of Munich (Munich, Germany)}
\affil[3]{Munich Center for Machine Learning (Munich, Germany)}
\date{\today}

\begin{document}
\maketitle

\begin{abstract}
Tabular foundation models such as TabPFN~\citep{hollmann2025accurate}, TabICL~\citep{qu2026tabiclv2}, ExaOneTabular~\citep{eo2026exaone}, and Nori~\citep{synthefy_2026_20710462} already produce full predictive distributions, yet prevailing regression benchmarks evaluate them almost exclusively via point-estimate metrics such as RMSE.
This overlooks the distributional information these models are designed to provide. %
We introduce \textbf{ScoringBench}\footnote{ScoringBench is available at \url{https://github.com/jonaslandsgesell/ScoringBench}.}, an open and extensible benchmark that evaluates tabular regression models under a comprehensive suite of proper scoring rules --- including (weighted) CRPS, energy scores and interval scores ---
alongside standard point metrics. ScoringBench covers 101 regression datasets from diverse domains, supports transparent community contributions via a git-based leaderboard, and provides two complementary ranking protocols:
an ordinal Dem\v{s}ar/autorank approach and a magnitude-preserving $z$-score ranking approach. %
Our evaluation of various model types, including (fine-tuned) tabular foundation models, gradient-boosted trees, and MLPs, shows that model rankings shift depending on the chosen scoring rule.
Models that perform well on point-estimate metrics can rank lower on probabilistic ones.
These findings demonstrate that evaluation metric selection is a core modeling decision and a domain-specific requirement for applications where certain types of errors carry higher consequences.
\end{abstract}

\section{Introduction}

The emergence of Prior-Data Fitted Networks (TabPFN)~\citep{hollmann2022tabpfn, hollmann2025accurate} has shifted tabular learning towards in-context learners~\citep{qu2026tabiclv2} that place attention on informative examples at inference time without model weight updates.
Both leading models (TabPFN and TabICLv2) perform \emph{discretized distributional regression} and produce full predictive distributions: TabPFN predicts a probability mass function over an adaptive grid, while TabICLv2 performs multi-quantile regression (approximating the CRPS training objective with a limited number of quantiles).

Despite this, prevailing tabular benchmarks (TabArena \citep{erickson2025tabarena}, TALENT \citep{liu2025talent}) evaluate models almost exclusively through point-estimate metrics such as RMSE or $R^2$, implicitly rewarding models that elicit a good conditional mean while ignoring the quality of the full predictive distribution.
This evaluation mismatch is not merely cosmetic.
\cite{gneiting2007strictly} proposed to use optimal score estimation as a general framework for estimation.
A scoring rule $S(\mathbb{P}, y)$ measures the quality of a probabilistic forecast $\mathbb{P}$ when a realization $y$ is observed and
is \emph{strictly proper} if the expected scoring rule is minimized if and only if the forecast $\mathbb{P}$ equals the true distribution.
A detailed overview of the scoring rules used in this article is given in \Cref{subsec:scoring_rules}.

\begin{figure}[h]
\centering
\includegraphics[width=0.5\textwidth]{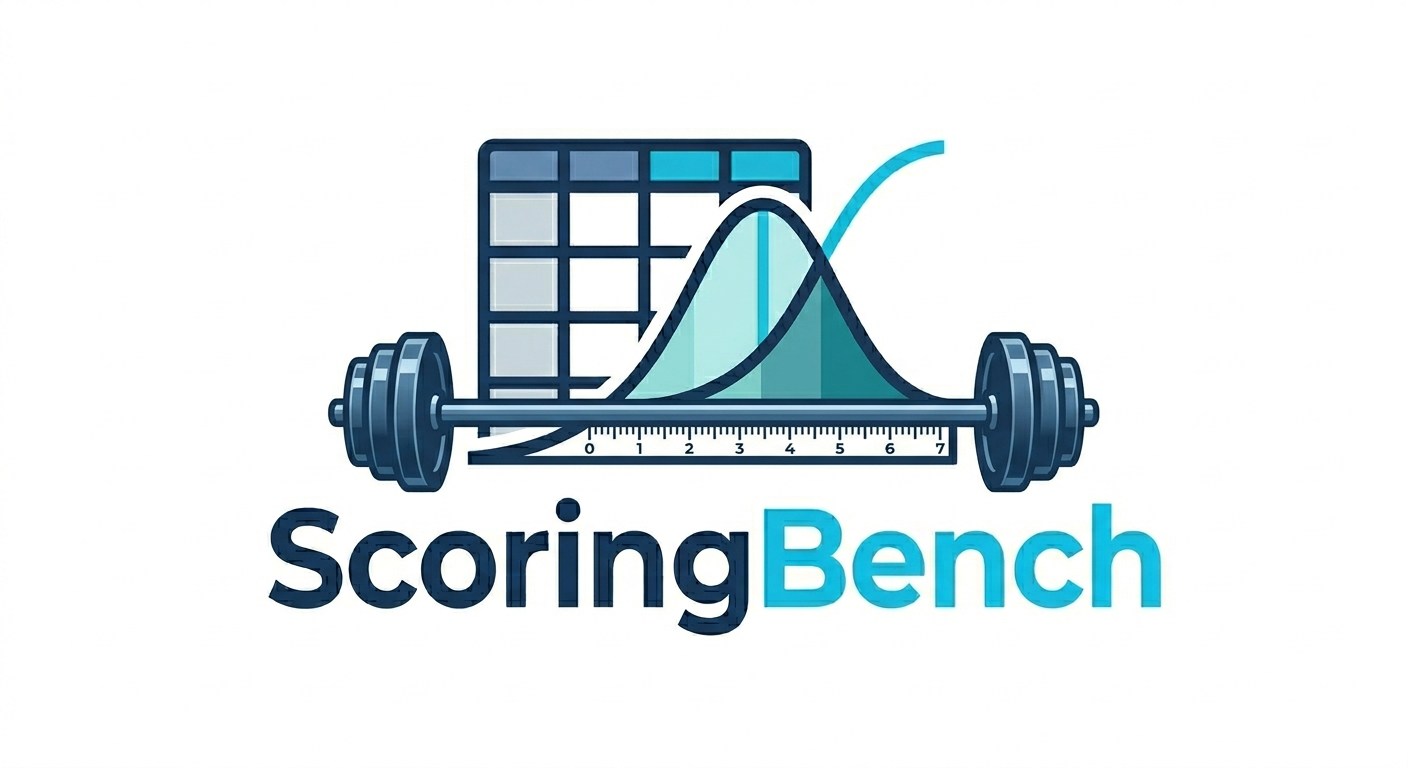}
\caption{ScoringBench: Evaluating tabular regression models with proper scoring rules.}
\label{fig:scoringbench}
\end{figure}

There are infinitely many proper scoring rules, each of which penalizes deviations in a particular way.
Thus different rules induce different inductive biases during finite-sample training,
even though they share the same population minimizer~\citep{landsgesell2026distributional, waghmare2025proper, buchweitz2026asymmetric}.
Consequently, different proper scoring rules yield different model rankings~\citep{merkle2013choosing, landsgesell2026distributional} and the choice of a scoring rule for training intuitively determines which aspects of the predictive distribution are learned first. 
For point estimation, \citet{gneiting2011making} argues that ``the common practice of requesting \emph{some} point forecast, and then evaluating the forecasters by using \emph{some} (set of) scoring function(s), is not a meaningful endeavor.''
Instead, effective forecasting requires clear directives, such as disclosing the evaluation metric \emph{ex ante}. 
The same argument applies to distributional regression: requesting a predictive distribution but not disclosing the scoring rule for evaluation risks weighting errors suboptimally for the undisclosed scoring rule (e.g. it is unclear if tail accuracy or central tendency shall be rewarded), especially in the finite-sample regime.

This inductive bias arising from the choice of scoring rule in the tabular distributional regression setting has been demonstrated in~\citet{landsgesell2026distributional}:
\begin{align*}
\left.
\begin{aligned}
\theta_1 &= \arg\min_{\theta} \hat{\mathbb{E}}_{(X_n, Y_n)_{n=1}^N} [S(\mathbb{P}_{\theta}(X_n), Y_n)], \\
\theta_2 &= \arg\min_{\theta} \hat{\mathbb{E}}_{(X_n, Y_n)_{n=1}^N} [S'(\mathbb{P}_{\theta}(X_n), Y_n)]
\end{aligned}
\right\}
\;\nRightarrow\;
\mathbb{P}_{\theta_1} = \mathbb{P}_{\theta_2}.
\end{align*}
Here, $\theta_1$ and $\theta_2$ are the parameter values obtained by minimizing the two scoring rules $S$ and $S'$; $\mathbb{P}_{\theta}(X_n)$ denotes the model's predictive distribution at input $X_n$, $Y_n$ is the observed outcome, 
and $\hat{\mathbb{E}}_{(X_n,Y_n)_{n=1}^N}$ is the sample average $\frac{1}{N} \sum_{n=1}^N(\cdot)$. 
The symbol $\nRightarrow$ indicates that, in finite samples, different empirical scoring rules can yield different fitted predictive mappings even when the scoring rules share the same population minimizer.

There is no universally optimal scoring rule: an appropriate choice depends on the application and the associated decision context.
As \citet{martin2022optimal} put it in the context of model selection, 
``one is forced to confront the fact that no such `true model' will be revealed, 
and that the criterion should be defined by a score that rewards the type of predictive accuracy that matters for the problem at hand'' 
-- an observation that applies equally to the choice of scoring rule for evaluating predictive distributions.
This is particularly important in high-risk applications, where different types of forecast errors may have strongly asymmetric consequences~\citep{gneiting2011making}: 
some such settings may require carefully designed custom proper scoring rules~\citep{oesterheld2020decision,johnstone2011tailored}, and even when the focus is on mean estimators, Bregman divergences can help induce desired residual structures~\citep{gneiting2011making}.

These differences have practical implications for model selection and deployment, motivating benchmarks that report a comprehensive suite of proper scoring rules alongside standard point metrics.

\textbf{ScoringBench} (\Cref{fig:scoringbench}) implements these ideas in an extensible, reproducible evaluation harness covering a broad set of scoring rules.
Our fine-tuning experiments illustrate how changing the training score affects predictive behavior.

\subsection{Contributions}

We contribute
\begin{itemize}
  \item ScoringBench (\url{https://github.com/jonaslandsgesell/ScoringBench}),
  an easy-to-install benchmark for distributional regression with minimal dependencies. 
  \item a git-based leaderboard (\url{https://scoringbench.com/}), where updates happen transparently with full traceability and reproducibility.
  \item a rigorous benchmark protocol using Autorank (following \citet{demsar2006statistical, Herbold2020}).
\end{itemize}

\section{Related Work}

\paragraph{Tabular Deep Learning.}
Prior-Data Fitted Networks (TabPFN)~\citep{hollmann2022tabpfn, hollmann2025accurate} have shifted tabular learning from the dominant paradigm of gradient-boosted decision trees toward transformer-based in-context learners.
Rather than updating model weights at fit time, PFNs perform in-context learning in a single forward pass. TabPFN predicts probability masses over an adaptive grid, while TabICLv2~\citep{qu2026tabiclv2}, ExaOneTabular~\citep{eo2026exaone} and Nori~\citep{synthefy_2026_20710462} estimates multiple conditional quantiles.

\paragraph{Probabilistic Forecasting.}
In a seminal review and theoretical development, \citet{gneiting2007strictly} surveyed proper scoring rules and their foundations, and discussed examples for probabilistic forecasts of densities, distributions, quantiles, and intervals.
\citet{Gneiting_sharpness_calibration2007} further formalized the goal of probabilistic forecasting as maximizing \emph{sharpness} subject to \emph{calibration}, providing the conceptual basis for the diagnostics reported in ScoringBench.
Importantly, while all strictly proper scoring rules share the same population-level minimizer, they induce different inductive biases during finite-sample training~\citep{waghmare2025proper}.
\citet{merkle2013choosing} demonstrated empirically that model rankings change substantially depending on which proper scoring rule is used (the Spearman correlation between Brier-score and log-score rankings in their study is only~$0.15$), and \citet{buchweitz2026asymmetric} showed that a broad class of proper scoring rules---including the logarithmic, CRPS, quadratic, and energy scores---penalize under- and overestimation asymmetrically.
The scoring rules computed in ScoringBench span a wide range of sensitivities: the CRPS~\citep{gneiting2007strictly} weights all quantile levels equally; the energy score~\citep{gneiting2007strictly} generalizes CRPS via the parameter~$\beta \in (0,2)$,
interpolating between median- and mean-optimal behavior~\citep{landsgesell2026distributional}.
The interval score~\citep{gneiting2007strictly} targets specific central prediction intervals; and the weighted CRPS~\citep{gneiting2011comparing} allows practitioners to emphasize particular regions of the predictive distribution.
Together, these rules enable a multifaceted assessment of forecast quality that no single metric can provide.
Additionally, we provide scoring rules which evaluate the density estimate (density power divergence scores and pseudo spherical scores, see appendix).

\paragraph{Benchmarking Methodology.}
Two prominent tabular benchmarks, TabArena~\citep{erickson2025tabarena} and TALENT~\citep{liu2025talent}, evaluate models exclusively on point-estimate metrics such as RMSE.
These metrics reward accurate conditional-mean predictions but are blind to the quality of the full predictive distribution.

\citet{landsgesell2026distributional} showed that fine-tuning realTabPFNv2.5~\citep{garg2025real, grinsztajn2025tabpfn} with a scoring rule different from the pretraining objective shifts the model's inductive bias and changes its rankings across models. 
Other recent work, including \cite{izbicki2026benchmarking}, has similarly highlighted the limitations of relying solely on point-estimate metrics; Izbicki et al. propose an additional loss termed ``conditional density estimation'', which can be viewed as a Brier-type loss for continuous outcomes \citep{rudemo1982empirical}. 
Building on this line of work \citep{landsgesell2026distributional}, ScoringBench offers a broad framework in which any admissible proper scoring rule can be used for evaluating the quality of predictive distributions. The benchmark establishes a reproducible, unified evaluation protocol for distributional regression.

\section{ScoringBench Setup}

\subsection{Benchmarking with Proper Scoring Rules}
We refer the reader to \cite{gneiting2007strictly, gneiting2011comparing} for a comprehensive introduction to proper scoring 
rules and \cite{landsgesell2026distributional} for a discussion in the tabular distributional regression setting.
All definitions follow~\citet{gneiting2007strictly,gneiting2011comparing} unless otherwise noted; 
full formulae, propriety classes, and implementation details are provided in the Appendix~\ref{subsec:scoring_rules} and Appendix~\ref{app:diagnostics}.

Since ScoringBench is not tied to a single downstream application, we report a representative suite of commonly used 
and complementary proper scoring rules alongside standard point metrics ($R^2$, RMSE, MAE): the \textbf{CRPS}, 
the $\beta$-\textbf{energy score} ($ES_\beta$, $\beta \in (0,2)$), the \textbf{continuous ranked logarithmic score} (CRLS), the \textbf{density power divergence} (DPD), 
the \textbf{interval score} (IS), the \textbf{quantile-weighted CRPS} (wCRPS), and the \textbf{CDE/quadratic (Brier-type) loss}. 
We use the loss convention throughout, so lower values indicate better forecasts. 
The scoring rule and its weighting should be specified according to the aspects of the predictive distribution 
--- e.g. tail behavior versus central tendency --- that are relevant to the downstream application; 
we refer to \citet{merkle2013choosing,oesterheld2020decision} for detailed discussions of the selection and design of scoring rules.
As a representative example, given a predictive CDF $F_{\mathbb{P}}$ and an observation $y$, the CRPS is
\begin{align*}
    \mathrm{CRPS}(\mathbb{P}, y) = \int_{-\infty}^{\infty} \big(F_{\mathbb{P}}(x) - \mathds{1}_{\{x \ge y\}}\big)^2\, \mathrm{d}x.
\end{align*}
When $\mathbb{P}$ collapses to a point mass at some $\hat y$, the CRPS reduces to the absolute error $|y-\hat y|$,
recovering point-estimate evaluation as a special case.
For univariate predictive distributions, the CRPS is furthermore the special case $\beta=1$ of the $\beta$-energy score.
Formulae for the remaining scoring rules, together with their natural domains of definition and further implementation details, 
are given in Appendix~\ref{subsec:scoring_rules}.

\subsection{Datasets}
\label{sec:datasets}
Many existing tabular benchmarks assemble datasets via semi-automated pipelines~\citep{gijsbers2024amlb, jiang2026omnitabbench, grinsztajn2022tree} 
or manual curation~\citep{erickson2025tabarena, grinsztajn2022tree}. Building on these efforts, ScoringBench curates a heterogeneous collection 
of regression problems spanning finance, physics, engineering, and environmental science.

Our dataset collection incorporates the three OpenML~\citep{bischl2025openml} regression
suites 269, 297, and~299 \citep{dheur2023large}.
We supplement these suites with datasets from three additional repositories: \textbf{PMLB}~\citep{olson2017pmlb}, \textbf{KEEL}~\citep{derrac2015keel}, and \textbf{TALENT}~\citep{liu2025talent}.
We drop duplicated datasets by filtering the names and also drop datasets that have only few unique target values (classification, not regression). 
Lightweight, uniform preprocessing is applied (e.g. numeric median imputation); see Appendix~\ref{app:datasets:preprocessing} for details on the imputation and preprocessing pipeline.
The complete list of datasets retained after deduplication and validation,
together with per-dataset details and selection criteria, is provided in Appendix~\ref{app:datasets}.

For each dataset, models (see Section~\ref{subsec:evaluated_models}) are evaluated using $5$-fold cross-validation with a fixed random seed. We limit each training fold to at most 2400 observations and each test fold to at most 600 observations. An ablation with regards to the number of observations is provided in Appendix~\ref{sec:ablation_dataset_size}.

\subsection{Evaluated Models}
\label{subsec:evaluated_models}

We evaluate a mix of in-context learners and conventional, non-pretrained baselines. 
At the time of writing, the in-context learners that support distributional prediction are \textbf{realTabPFNv2.5} (and several fine-tuned variants of realTabPFNv2.5), \textbf{TabPFNv2.6}, \textbf{TabPFNv3}~\citep{grinsztajn2026tabpfn}, \textbf{TabICLv2}~\citep{qu2026tabiclv2},
\textbf{ExaOneTabular}~\citep{eo2026exaone}, and Nori~\citep{synthefy_2026_20710462}.
We expect this list to grow as further models add distributional prediction.\footnote{This is precisely the advantage of maintaining ScoringBench as a living benchmark: rather than being fixed at publication time, it continues to track newly released tabular foundation models on the git-based leaderboard. Since this manuscript was written, \textbf{TabDPT}~\citep{hosseinzadeh2026tabdpt}, \textbf{Mitra-v2}~\citep{tao2026mitra}, \textbf{LimiX-2}~\citep{zhang2026limix}, \textbf{Causilo}~\citep{cho2026causilo}, and \textbf{Xiaomi-TabLDM}~\citep{wang2026tabldm} have already been added; up-to-date results for these and future models are available at \url{https://scoringbench.com/}.}
Our non-pretrained baselines include \textbf{XGBoost vector-leaf} models adapted for distributional regression,
\textbf{XGBoostLLS} \citep{marz2019xgboostlss}, and \textbf{CatBoost} \citep{prokhorenkova2018catboost}.
We also use \textbf{CREPES} \citep{bostrom2024} to construct conformal predictive distributions from XGBoost,
CatBoost, and TabICLv2 predictions.
We also evaluate \textbf{TabM} \citep{gorishniy2024tabm}, a compact multilayer perceptron that leverages parameter-efficient ensembling, and \textbf{RealMLP} \citep{holzmuller2024better}, 
an MLP architecture with extensive hyperparameter tuning designed for robust out-of-the-box performance on diverse tabular tasks.

\subsection{Implementation}
\label{subsec:implementation}

The models under consideration originally represent their predictive distributions in different forms, 
including histograms, quantiles, samples, and probability density functions.
For evaluation in ScoringBench, model-specific wrappers convert these outputs into a common histogram representation.
For a given observation $i$, this representation consists of bins $I_{ik}=[a_{ik}, b_{ik}]$ and associated probability masses $p_{ik}$, with $\sum_k p_{ik}=1$.
For density-based scoring rules and models without native histogram representations, we adapt \citet{izbicki2026benchmarking} to enable reliable conditional density estimation:
we monotonically interpolate the cumulative distribution function (CDF),
create a histogram with equal bin widths, and compute the probability mass in each bin by evaluating the interpolated CDF at the bin edges ($p_{ik}=F(b_{ik})-F(a_{ik})$) which is effectively a smoothing scheme.

Within each bin, the probability is assumed to be distributed uniformly, resulting in the piecewise-constant density
$f_i(x)= p_{ik} / (b_{ik}-a_{ik})$, $x \in I_{ik}$.
This unified representation allows the required quantities to be evaluated analytically within each bin.
For example, for positive-width bins, the first term of the CRPS is then computed as
\begin{align*}
\mathbb{E}|X-y_i|=\sum_k p_{ik}\frac{1}{b_{ik}-a_{ik}}
\int_{a_{ik}}^{b_{ik}} |x-y_i|\,\mathrm{d}x,
\end{align*}
which can be computed analytically.
Other scoring rules are calculated analogously.

\subsection{Ranking}
\label{sec:ranking}

We use two complementary ranking strategies; full details are in Appendix~\ref{app:details_ranking_methodology}. 
The benchmark produces scores $s_{m,d,f,p}$ for each tuple
\begin{align*}
    (\text{model } m,~\text{dataset } d,~\text{fold } f,~\text{metric } p),
\end{align*}
across $M$ models, $D$ datasets, $F$ folds, and $P$ metrics (in total $M \cdot D \cdot F \cdot P$ observations). 
Before ranking, we collapse across folds by averaging,
\begin{align*}
  \bar{s}_{m,d,p} \;=\; \frac{1}{F}\sum_{f=1}^{F} s_{m,d,f,p}\,,
\end{align*}
reducing the data to per-(model, dataset, metric) scores. 
This step is essential to avoid \emph{pseudoreplication}: fold-level scores within a dataset are correlated, so treating them as independent observations would inflate the effective sample size and distort subsequent tests~\citep{lazic2010problem}.
After fold aggregation the data has shape $M \times D \times P$, with entries $\bar{s}_{m,d,p}$ denoting the fold-averaged score of model $m$ on dataset $d$ for metric $p$.

The two ranking strategies differ in how they aggregate these per-dataset summaries across datasets:

\paragraph{Dem\v{s}ar--autorank approach (Autorank).} Following Dem\v{s}ar~\citep{demsar2006statistical} and the \texttt{autorank} library~\citep{Herbold2020}, models are ranked within each dataset (best = rank ~1),
mean ranks are computed across datasets, and a non-parametric Friedman omnibus test evaluates whether mean ranks differ across models. 
If the omnibus test is significant, Nemenyi post-hoc comparisons (visualized via a critical-difference diagram) identify which model pairs differ; 
the Nemenyi test is automatically selected by autorank as the appropriate non-parametric post-hoc procedure for this Demšar-based approach.
Autorank also reports Akinshin's~$\gamma$ as a non-parametric analogue of Cohen's~$d$.\footnote{\url{https://aakinshin.net/posts/nonparametric-effect-size/}}
Thresholds on the absolute effect size $|\gamma|$ used to categorize magnitudes in the Autorank library follow the common practice for Cohen's $d$: $|\gamma|<0.2$ is negligible; $0.2\le|\gamma|<0.5$ is small; $0.5\le|\gamma|<0.8$ is medium; $|\gamma|\ge0.8$ is large.
This ordinal approach is robust to outliers and to heteroskedastic score scales, but it discards magnitude information by relying on ranks. 
Some more details are provided in \Cref{subsec:autorank}.
For results see the Autorank-based heatmap in \Cref{fig:heatmap_autorank}. 

\paragraph{Per-dataset standardized performance approach ($z$-score ranking).} To preserve magnitude information while controlling for dataset difficulty, we standardize the fold-averaged scores within each dataset (compute $z$-scores across the $M$ models, negating the sign for metrics where smaller is better), 
then average these standardized scores across datasets and rank models by their mean standardized performance. 
The full details are given in \Cref{subsec:app:zscore}.
This produces magnitude-preserving, comparable scores across datasets that account for differing score scales and difficulties. See the effect-size heatmap in \Cref{fig:heatmap_effect_size}. \\

We refer to the Dem\v{s}ar--autorank procedure as \textbf{Autorank} and the per-dataset standardized performance procedure as \textbf{$z$-score ranking}; subsequent figures and tables use these labels.
The two approaches are complementary (ordinal significance vs.\ magnitude-aware ranking); for implementation details see Appendix~\ref{app:details_ranking_methodology}. 
Since both produce highly correlated rankings across metrics (mean Pearson $r \approx 0.97$; see \Cref{subsec:ranking_stability}), we report only the
\textbf{Autorank}-based leaderboards in the main text. Full $z$-score rankings and per-metric correlation statistics are available on the live leaderboard and may be relevant for readers interested in magnitude-aware differences between models.

\section{Results of Benchmark 2026}

The reported results represent a snapshot from 8th September 2026.

Our website 
\begin{center}
\url{https://scoringbench.com/}
\end{center}
reports the live leaderboard with all models and their respective rankings. 
Hyperparameters of the models are documented in the git repository. 
The leaderboard is updated with every new pull request, and all changes are tracked in the git history for full reproducibility, transparency and agility.

\begin{figure}[ht]
\centering
\begin{subfigure}[b]{0.48\textwidth}
  \centering
  \includegraphics[width=\textwidth]{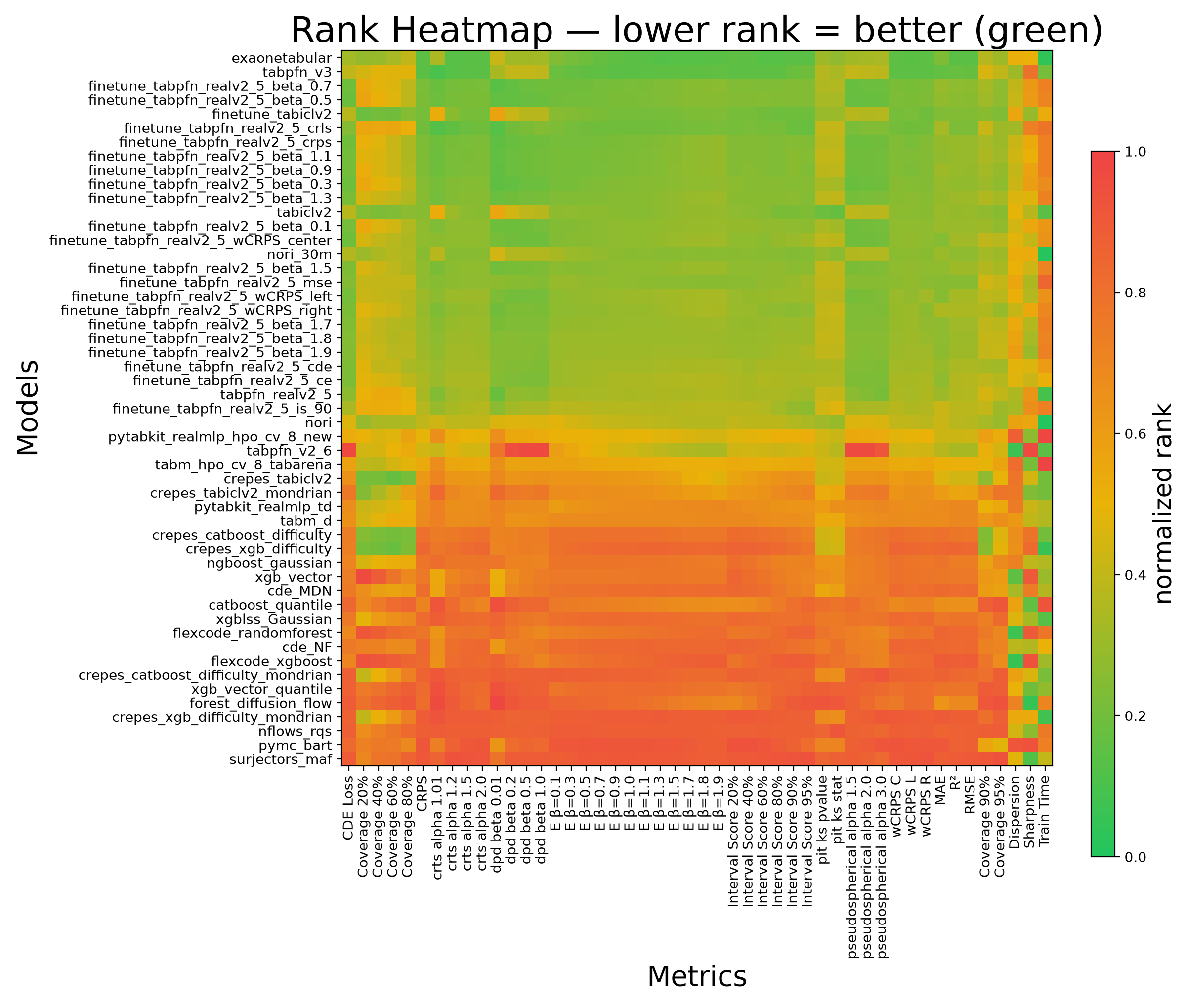}
  \caption{Autorank}
  \label{fig:heatmap_autorank}
\end{subfigure}\hfill
\begin{subfigure}[b]{0.48\textwidth}
  \centering
  \includegraphics[width=\textwidth]{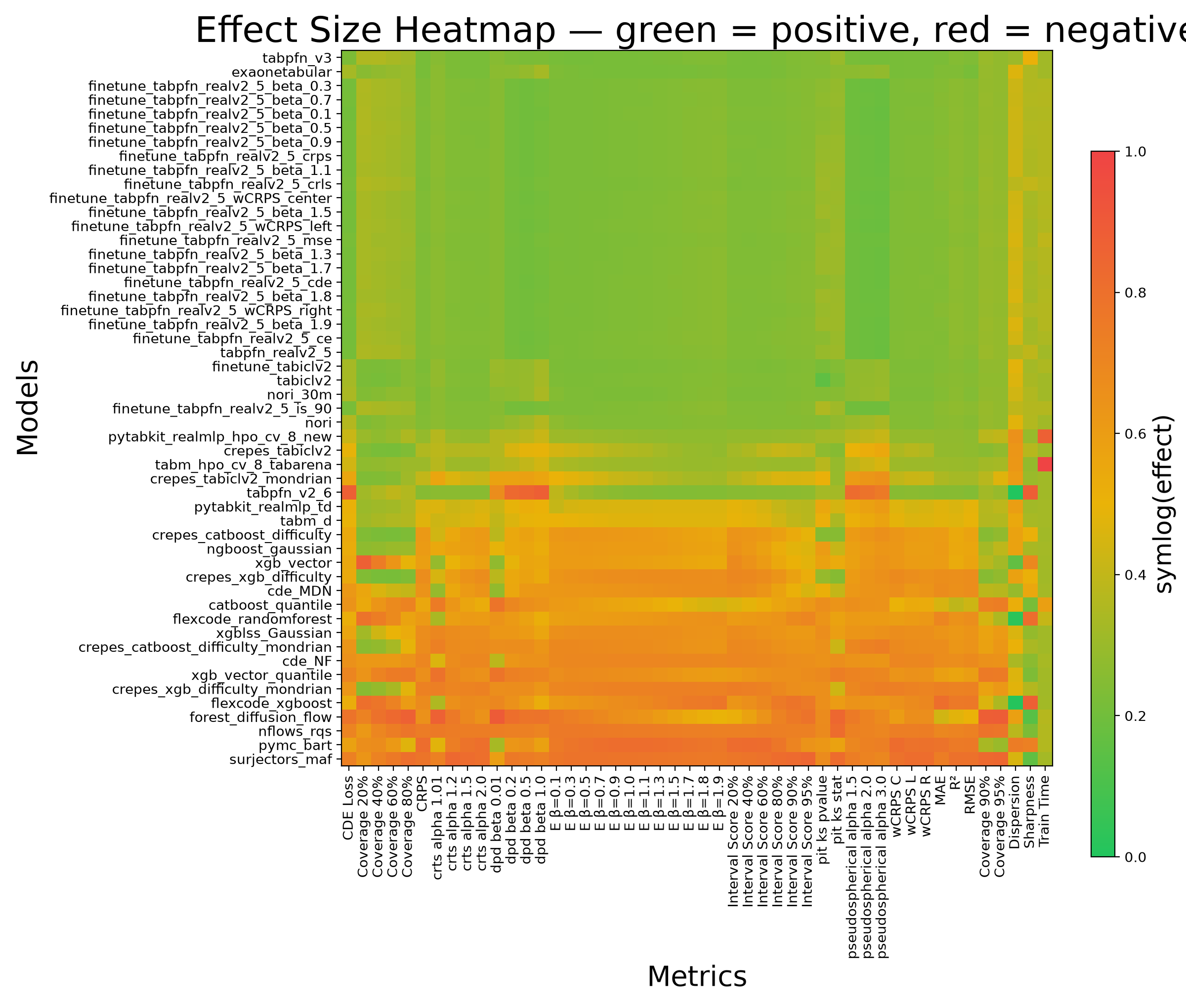}
  \caption{$z$-score ranking}
  \label{fig:heatmap_effect_size}
\end{subfigure}
\caption{Ranking heatmaps summarizing the performance of different models on different scoring rules.
}
\label{fig:heatmaps}
\end{figure}

The results confirm the central finding of~\citet{landsgesell2026distributional}: 
the ranking of different models depends on the choice of scoring rule, and reporting only point-estimate metrics risks obscuring these differences.

\Cref{tab:leaderboard_crps} reports Autorank-based rankings for CRPS (lower is better); column headers are defined in the table caption.
The leading models are \texttt{exaonetabular} and \texttt{tabpfn\_v3}~\citep{grinsztajn2026tabpfn} (mean ranks $8.72$ and $8.73$), followed by
\texttt{finetune\_tabpfn\_realv2\_5\_crls} (mean rank $11.08$).
Several fine-tuned TabPFN variants also rank ahead of the unfine-tuned \texttt{tabpfn\_realv2\_5} (mean rank $18.21$).
While Akinshin's $\gamma$ values relative to the top-ranked model remain in the ``negligible'' range for these models, the rank ordering is
consistent across datasets and robust to dataset size (see \Cref{sec:ablation_dataset_size}), suggesting that fine-tuning induces 
subtle but stable shifts in predictive behavior. Magnitude categories use $|\gamma|$ and follow the common practice for Cohen's $d$ according to~\cite{romano2006appropriate}.

\begin{figure}
\centering
\includegraphics[width=0.8\textwidth]{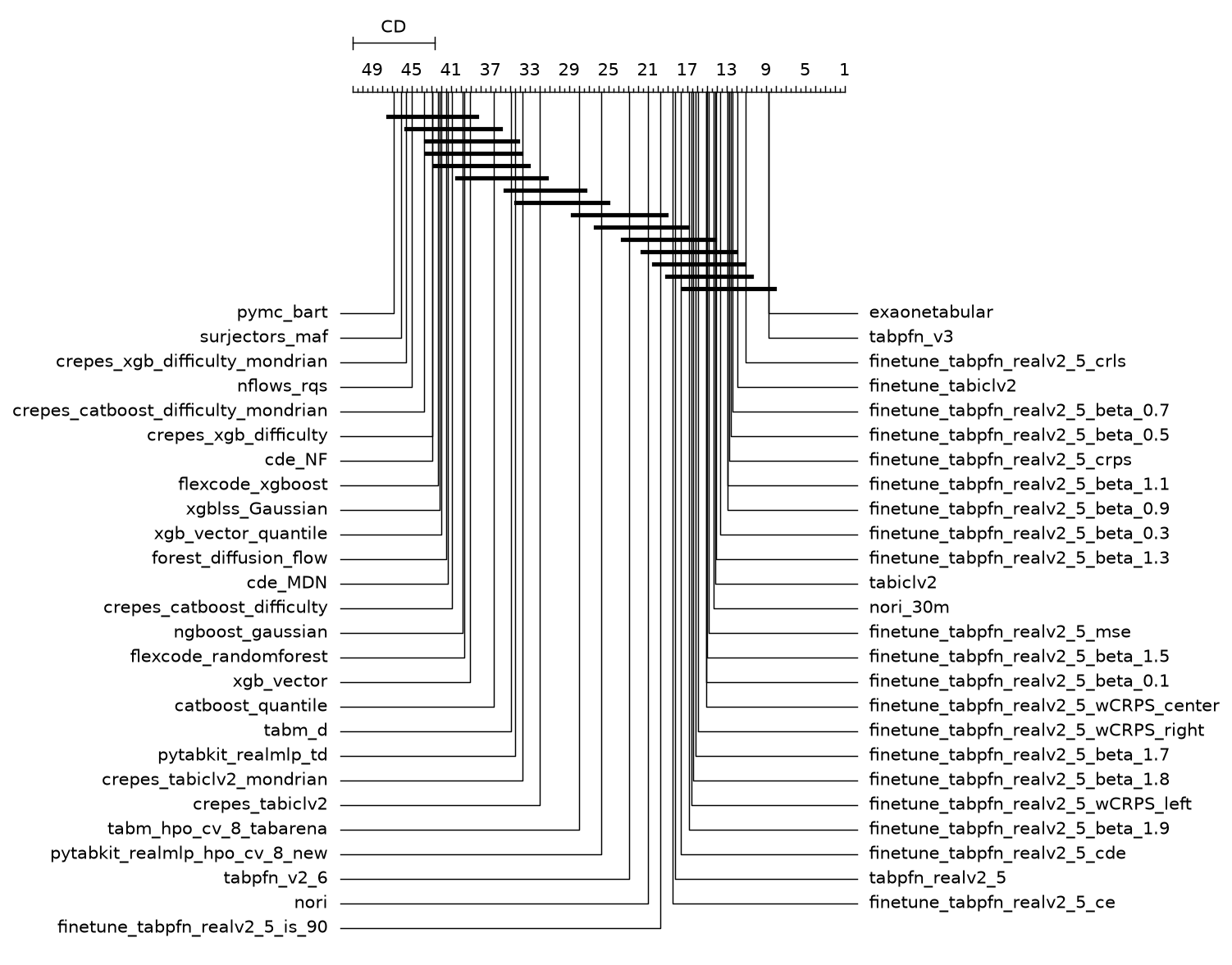}
\caption{Critical Difference (CD) diagram for CRPS: Models are positioned by autorank on a horizontal axis according to their mean rank $\bar{r}_{m,p}$ (see Eq.~\eqref{eq:mean_rank}),
and those whose mean-rank difference does not exceed $\mathrm{CD}$ are connected by a horizontal bar, indicating no statistically significant difference in performance at the $\alpha=0.05$ level. Critical difference (CD) is defined in Eq.~\eqref{eq:CD_value} \citep{demsar2006statistical}.}
\label{fig:cd_diagram_crps}
\end{figure}

The critical-difference diagram in \Cref{fig:cd_diagram_crps} shows that tabular foundation models provide substantial improvements over traditional baselines: 
the top (fine-tuned) variants form a cluster that is statistically significantly better than the baselines such as \texttt{xgb\_vector} and \texttt{xgblss\_Gaussian}.


While the importance of aligning the scoring rule used for training with application-specific utility is well-established among leading statisticians --- most notably popularized by~\cite{gneiting2007strictly,gneiting2011comparing,gneiting2011making} and colleagues --- its practical implementation in terms of optimal score estimation remains frequently overlooked. 

In general, we observe---consistent with \citep{landsgesell2026distributional}---that the ranking of models depends on the specific scoring rule used to penalize errors (or the chosen scoring function for point estimates~\citep{gneiting2011making}).
This can also be seen by comparing the different leaderboards \Cref{tab:leaderboard_crps} and \Cref{tab:leaderboard_rmse},
see e.g.\ the entries for TabPFNv2.6.
In view of \Cref{fig:heatmaps},
we observe correlations between different scoring rules,
in particular between scoring rules belonging to the same scoring rule family.
For an investigation of this effect, 
we point to \cite[Figure 10]{landsgesell2026distributional}.

We also observe that wrapping TabICLv2 in conformal methods like CREPES does not improve the interval score and other proper scoring rules (see \Cref{tab:leaderboard_is90}).
Proper scoring rules reward calibration \emph{and} sharpness simultaneously~\citep{gneiting2007strictly}, so for a model such as TabICLv2, which is trained on CRPS, such a wrapper compromises the distributional quality it was designed to achieve (also absolute marginal coverage is not necessarily improved by conformal methods when the absolute value is taken before fold averaging).

\FloatBarrier
\section{Conclusion}
\label{sec:conclusion}
 
ScoringBench is an open, extensible benchmark that closes the evaluation gap between the full predictive distributions produced by tabular foundation models and the point estimates measured by prevailing benchmarks. 
By reporting a comprehensive suite of proper scoring rules alongside standard metrics, ScoringBench 
enables a more faithful assessment of probabilistic forecast quality, especially for applications where the cost of errors is asymmetric.

Our results support three main findings. First, model rankings depend on the chosen scoring rule: point-estimate metrics (scoring functions) such as RMSE can paint a different picture than proper scoring rules that evaluate full predictive distributions. Second, fine-tuning realTabPFN with
alternative scoring rules tends to produce small effect sizes (Akinshin's $\gamma$ often in the negligible range) but yields consistent directional shifts across datasets and metrics. Third, pretrained in-context models achieve substantial improvements over gradient-boosted baselines on probabilistic metrics, indicating genuine recent progress in distributional regression.

We therefore recommend that benchmarks include a diverse suite of proper scoring rules and present both rank-based and magnitude-aware summaries. For high-risk applications, practitioners should select~\citep{gneiting2011making} or design~\citep{merkle2013choosing, oesterheld2020decision} scoring rules that encode domain-specific utility. For example, weighted CRPS suits settings where tail errors are disproportionately costly.

A natural next step is multivariate distributional regression, where predictive quality depends on the dependence structure between targets and not only on the marginals.
Multivariate proper scoring rules such as the energy score~\citep{gneiting2007strictly} and the variogram score~\citep{scheuerer2015variogram} assess this joint structure, but their discriminative power weakens as the dimension grows~\citep{pinson2013discrimination}.
The choice of scoring rule also matters beyond regression: training language models with strictly proper scoring rules other than the logarithmic score can improve generation quality~\citep{shao2024language}.

We see ScoringBench as a living resource and welcome community contributions of new datasets, models, and scoring rules via pull request. 
The code and contribution guide are available at \url{https://github.com/jonaslandsgesell/ScoringBench}. The live leaderboard can be found at \url{https://scoringbench.com/}.

\paragraph{Limitations}
ScoringBench currently takes into account only scalar valued output datasets.
However this does not seem to be a structural limitation, since there are also proper scoring rules for vector valued outputs.

\subsection*{AI use statement}

In this work, we used generative AI tools for
providing feedback on experiments,
implementing methods, assisting with translation, cleaning and reformatting datasets,
supporting qualitative and thematic data analysis.
We have not used generative AI tools for
generating synthetic data sets,
helping develop theoretical models or conceptual frameworks,
proposing or refining hypotheses,
interpreting results.
``Formulate mathematical claims, provide critical ingredients for proving mathematical claims,
assist in the writing of proofs''
are not applicable to this work.
Additionally, we used generative AI tools for
creating or modifying scientific figures,
suggesting experimental parameters,
creating and editing software code,
drafting parts of a research paper,
summarizing or analysing existing literature,
sourcing/searching for information,
editing a research paper to improve readability,
identifying relevant literature.

We have reviewed all AI-assisted work.
LLM-generated code was verified and tested by all three authors,
all papers that were found by AI were reviewed, understood and approved each by at least two authors.
the drafted text was reviewed, manually edited and approved by all three authors.
We take responsibility for the final content of this work,
including text, claims or artifacts produced with the aid of generative AI.

\subsection*{Ethics statement}

ScoringBench evaluates probabilistic regression models to improve understanding of how their performance depends
on the choice of scoring rule.
We do not identify immediate ethical risks arising from the benchmark itself.
However, benchmark performance alone does not establish that a model is suitable for a particular application,
especially where errors may affect people. Users should also assess fairness, privacy,
and the consequences of prediction errors in their intended setting.

\subsection*{Reproducibility Statement.}

The ScoringBench code is available at the link provided in the paper.
The repository README explains how to rerun the benchmark and reproduce the reported results.
We describe the computational resources used for the experiments in the paper,
and the ScoringBench website is publicly available.

\bibliographystyle{unsrtnat}
\bibliography{references}

\FloatBarrier
\include{appendix.tex}

\end{document}

%% file: appendix.tex
\FloatBarrier

\appendix

\section{Scoring Rules and Diagnostic Metrics}
\label{app:metrics}

\subsection{Scoring rules}
\label{subsec:scoring_rules}

This appendix collects concise definitions of the scoring rules and diagnostic metrics used in ScoringBench.  
Full implementation notes are available in the repository.
We use the loss convention throughout, so lower values indicate better forecasts.

In the following, let $F_{\mathbb{P}}$ be the cumulative distribution function (CDF) of the probability distribution $\mathbb{P}$,
let $f_\mathbb{P}$ be the predictive density of $\mathbb{P}$ and let $y$ be an observation.
Before giving the definitions of the most important scoring rules,
we recall that (strict) propriety is relative to the class of admissible predictive distributions \citep{gneiting2021mcmc, gneiting2007strictly}.
For example, the logarithmic score is strictly proper with respect to the class $\mathcal{L}_1$
of absolutely continuous probability measures,
while the CRPS is strictly proper with respect to the class $\mathcal{M}_1$ of Borel probability measures
that have finite first moment \citep{gneiting2021mcmc, gneiting2007strictly}.
For details on the scoring rules and their properties, we refer to \citep{gneiting2007strictly}.
This distinction matters in practice: a scoring rule that is inapplicable, or numerically unstable
can favor or penalize models in an inappropriate way.
Therefore, when comparing models across scoring rules, as ScoringBench does, it is therefore important to verify
that the kind of output prediction of the model is valid for the scoring rule used.
For density-based scores, we smooth potential atoms (e.g., from tied quantiles) by monotone CDF interpolation and equal-width binning~\citep{izbicki2026benchmarking}.
CRPS and energy scores use native forecasts without this smoothing.

\paragraph{Continuous Ranked Probability Score (CRPS).}
The CRPS is given as
\begin{align*}
    \text{CRPS}(\mathbb{P}, y) = \int_{-\infty}^{\infty} (F_{\mathbb{P}}(x) - \mathds{1}_{\{x \geq y\}})^2 \, \mathrm{d}x.
\end{align*}
For univariate probability distributions,
the CRPS is a special case of the $\beta$-energy score $ES_{\beta=1}$.

\paragraph{$\beta$-energy score.}
For any $\beta \in (0,2)$, the $\beta$-energy score is given as
\begin{align*}
ES_{\beta}(\mathbb{P}, y) = \mathbb{E}_{X \sim \mathbb{P}} \|X - y\|^\beta - \tfrac{1}{2}\, \mathbb{E}_{X, X' \sim \mathbb{P}} \|X - X'\|^\beta,
\end{align*}
where $X, X' \sim \mathbb{P}$ are independent draws from $\mathbb{P}$. 
Smaller values of $\beta$ reduce the relative influence of large errors,
while values of $\beta$ closer to $2$ emphasize large deviations more strongly.
In the limiting case $\beta \rightarrow 2$,
the MSE is obtained.

\paragraph{Continuous Ranked Logarithmic Score (CRLS).}

Following \citet{juutilainen2012exceedance}, the CRLS is given as:
\begin{align*}
    \mathrm{CRLS}(\mathbb{P}, y) = -\int_{-\infty}^{\infty} \log |F_{\mathbb{P}}(x) + \mathds{1}_{\{y \le x\}} - 1| ~ \mathrm{d}x,
\end{align*}

\paragraph{Density Power Divergence (DPD) scoring rule.}
Following \citet{ghosh2013robust}, the DPD is given as:
\begin{align*}
    DPD_\beta(\mathbb{P}, y) = \int_{-\infty}^{\infty}  f_\mathbb{P}(t)^{1+\beta} ~ \mathrm{d}t - \left(1 + \frac{1}{\beta}\right) f_\mathbb{P}(y)^\beta, \qquad \beta>0.
\end{align*}
In the limiting case $\beta\to 0$ this recovers the log score up to an additive constant:
\begin{align*}
    DPD_0(\mathbb{P}, y) = -\log f_\mathbb{P}(y) + \text{const.}
\end{align*}
Note for $\beta=1$ this corresponds to the CDE loss, which is a continuous generalization of the Brier score.

\paragraph{Interval Score.}
The Interval Score (IS) \citep{gneiting2007strictly} evaluates the $(1{-}\alpha)$ central prediction interval:
\begin{align*}
    S_{\alpha}(\mathbb{P}, y) = (u-l) + \frac{2}{\alpha}(l-y)\mathds{1}_{\{y < l\}} + \frac{2}{\alpha}(y-u)\mathds{1}_{\{y > u\}},
\end{align*}
where $[l, u]$ are the $\alpha/2$ and $1{-}\alpha/2$ quantiles of the CDF $F_\mathbb{P}$. 

\paragraph{Quantile Weighted CRPS (wCRPS).}
The weighted CRPS can be written as an integral of the pinball loss \(\rho_\alpha\) over quantile levels \(\alpha\in[0,1]\). Using a weight function \(w(\alpha)\) to emphasise regions of interest, a convenient representation is
\begin{align*}
    L_{\mathrm{wCRPS}}(\mathbb{P},y) 
    = 2\int_{0}^{1} w(\alpha)\, \rho_{\alpha}\bigl(y - F_\mathbb{P}^{-1}(\alpha)\bigr) ~ \mathrm{d}\alpha,
\end{align*}
where the pinball loss is defined by
\begin{align*}
    \rho_{\alpha}(u) = (\alpha - \mathds{1}_{\{u < 0\}})\,u.
\end{align*}
The weight functions for the left, center, and right cases are given, respectively, 
by $w(\alpha) = (1-\alpha)^2$, $w(\alpha) = \alpha(1-\alpha)$, $w(\alpha) = \alpha^2$.

\paragraph{The CDE Loss as Continuous Brier Score}

The Conditional Density Estimation (CDE) loss (sometimes quadratic score) \citep{rudemo1982empirical, izbicki2016nonparametric} generalises the Brier Score \citep{gneiting2007strictly} to continuous outcomes by minimising the $L^2$ distance between the true density $f$ and an estimator $\hat f$:
\begin{align*}
        L(f,\hat f) 
        = \int \bigl(f(z) - \hat f(z)\bigr)^2 ~ \mathrm{d}z.
\end{align*}
Dropping the estimator-independent term $\int f^2(z) ~ \mathrm{d}z$ yields the modified CDE loss functional used for estimation:
\begin{align*}
        \mathcal{L}_{\mathrm{CDE}}(\hat f, y) 
        = \int \hat f^2(z) ~ \mathrm{d}z - 2\,\hat f(y).
\end{align*}
This expression is directly analogous to the discrete Brier loss $\sum_k p_k^2 - 2p_y$, 
where the integral of the squared density replaces the sum of squared probabilities and the density value at the observation replaces the ``hit'' probability. 
Importantly, numerical stability of $\mathcal{L}_{\mathrm{CDE}}$ depends on well-defined supports and strictly monotonic bin boundaries (for example, sorted quantiles in discretised estimators). 
Because $\mathcal{L}_{\mathrm{CDE}}$ is strictly proper, it incentivises calibrated predictive densities.

\subsection{Diagnostic Metrics}
\label{app:diagnostics}
\paragraph{Sharpness.}
Sharpness measures the concentration of the predictive distribution.
Following \citet{tran2020methods} we compute sharpness from the model's estimated predictive CDFs \(\widehat{F}_n\) for each test point \(n\) as the average predicted standard deviation:
\begin{align*}
    \mathrm{Sharp}(\widehat{F}) = \frac{1}{N}\sum_{n=1}^{N}\sqrt{\mathrm{var}(\widehat{F}_n)}
    = \frac{1}{N}\sum_{n=1}^{N}\sqrt{\mathbb{E}_{\widehat{F}_n}[X^2] - \bigl(\mathbb{E}_{\widehat{F}_n}[X]\bigr)^2}.
\end{align*}
Here, \(\widehat{F}_n\) denotes the model's estimated predictive CDF for test sample \(n\), and we write \(X\sim\widehat{F}_n\). Define \(\sigma_n = \sqrt{\mathrm{var}(\widehat{F}_n)}\) and the mean predicted uncertainty \(\bar{\sigma}=\frac{1}{N}\sum_{n=1}^N\sigma_n\).
Sharpness is not a proper scoring rule; it should be considered jointly with other scoring rules. A well-calibrated model with low sharpness (small predicted standard deviations) provides more actionable uncertainty 
estimates than one that is calibrated but diffuse \citep{tran2020methods}.

\paragraph{Dispersion.}
Dispersion measures heterogeneity in the model's predicted uncertainties across the test set. Using the notation above, let \(\sigma_n=\sqrt{\mathrm{var}(\widehat{F}_n)}\) be the predicted standard deviation for sample \(n\). Then dispersion is the cross-sample standard deviation of these predicted uncertainties:
\begin{align*}
    \mathrm{Disp}(\widehat{F}) 
    = \sqrt{\frac{1}{N-1}\sum_{n=1}^N (\sigma_n - \bar{\sigma})^2},
\end{align*}
where \(\bar{\sigma}=\frac{1}{N}\sum_{n=1}^N\sigma_n\). Dispersion is not a proper scoring rule; it answers the question ``how much does the model's confidence vary across samples?''.
A low-dispersion model assigns similar uncertainty estimates to all instances, while a high-dispersion model distinguishes some instances as more predictable than others. Dispersion complements sharpness: two models with the same average sharpness may differ substantially in dispersion.

Higher dispersion is favorable only when true underlying errors are heteroscedastic (i.e., when prediction difficulty genuinely varies across samples). If errors are homoscedastic, a model with high dispersion wastes capacity on spurious confidence variation.
We report dispersion alongside sharpness and other metrics for completeness, but treat it as a secondary diagnostic tool (not a primary optimization target) that is only meaningful for datasets with heteroscedastic errors.

\paragraph{Coverage.}
For a nominal $(1-\alpha)$ prediction interval $[l, u]$ extracted from the predictive CDF, empirical coverage is the fraction of test points falling inside the interval:
\begin{align*}
  \mathrm{Cov}_{1-\alpha} = \frac{1}{N}\sum_{n=1}^N \mathds{1}_{l_n \le y_n \le u_n}.
\end{align*}
Here, for the $n$-th test sample, $[l_n,u_n]$ denotes the nominal $(1-\alpha)$ predictive interval (lower and upper bounds) extracted from the model's predictive CDF for that sample, 
and $y_n$ is the observed realization of the target.
We report coverage at the 90\% level ($\mathrm{Cov}_{90}$, target 0.90) and the 95\% level ($\mathrm{Cov}_{95}$, target 0.95). 
Concretely, for $\mathrm{Cov}_{90}$ we set $\alpha=0.10$ (nominal level $1-\alpha=0.90$) and for $\mathrm{Cov}_{95}$ we set $\alpha=0.05$ (nominal level $1-\alpha=0.95$); 
$\alpha$ denotes the total tail probability excluded from a central $(1-\alpha)$ interval $[l_n,u_n]$.
A well-calibrated model achieves coverage close to the nominal level; the interval score (IS) jointly penalizes over-wide intervals and coverage shortfalls.

\paragraph{PIT Kolmogorov--Smirnov test.}
The probability integral transform (PIT) assesses the calibration of the full predictive distribution rather than just an interval or a single moment \citep{dawid1984present, diebold1998}.
For test sample $n$ with predictive CDF $\widehat{F}_n$ and observed outcome $y_n$,
the PIT value is
\begin{align*}
    p_n = \widehat{F}_n(y_n).
\end{align*}
If the predictive distributions are ideal (i.e., $y_n$ is truly drawn from $\widehat{F}_n$ for every $n$),
the PIT values $\{p_n\}_{n=1}^N$ are i.i.d.\ $\mathrm{Uniform}(0,1)$,
regardless of the shape of $\widehat{F}_n$ itself \citep{dawid1984present}.
We test this null hypothesis with a one-sample Kolmogorov--Smirnov test against $\mathrm{Uniform}(0,1)$,
reporting the test statistic
\begin{align*}
    \mathrm{PIT\text{-}KS} = \sup_{p\in[0,1]} \bigl| \widehat{G}_N(p) - p \bigr|,
\end{align*}
where $\widehat{G}_N$ is the empirical CDF of $\{p_n\}_{n=1}^N$,
together with the corresponding p-value $\mathrm{PIT\text{-}KS}_{\text{pval}}$.
A PIT-KS statistic near zero (equivalently, a large p-value) indicates that the predicted distributions are statistically indistinguishable from calibrated;
a small p-value indicates systematic miscalibration (e.g., predictive distributions that are too narrow, too wide, or biased).
Following the histogram-based prediction representation,
PIT values falling within a bin are obtained by linear interpolation of the piecewise-constant predictive density,
and PIT values landing exactly on a probability atom (zero-width bin) are assigned the mid-CDF value $0.5$,
the standard non-randomized convention for discrete or tied predictive mass \citep{czado2009predictive}.
PIT-KS is not a proper scoring rule: it is a pure calibration diagnostic and,
unlike sharpness or dispersion, is agnostic to how concentrated the predictive distribution is;
a model can be perfectly calibrated (PIT-KS $\approx 0$) while still being uninformative (low sharpness).
We therefore report PIT-KS jointly with sharpness, dispersion, and coverage
for diffent aspects of predictive performance.

\section{Benchmark Datasets}
\label{app:datasets}


We draw candidate datasets from four complementary public
repositories (Section~\ref{app:datasets:sources}), apply a multi-level
deduplication procedure to remove redundant entries
(Section~\ref{app:datasets:dedup}), and filter the remaining candidates to
retain only well-formed regression tasks
(Section~\ref{app:datasets:validation}). All datasets are available through
public APIs (OpenML, PMLB GitHub, KEEL) or direct download URLs without
requiring authentication, ensuring full reproducibility. Our guiding
principles follow recent calls for higher-quality tabular
benchmarks~\citep{erickson2025tabarena, grinsztajn2022tree}: we prioritize
diversity across domains and feature dimensionalities, favour datasets derived
from genuine data-collection processes, and deduplicate aggressively to avoid
inflating the effective benchmark size with near-identical tasks.

\subsection{Data Sources}
\label{app:datasets:sources}

Candidate datasets are drawn from four complementary repositories:

\begin{enumerate}
    \item \textbf{OpenML benchmark suites.}
    We include three curated regression suites hosted on
    OpenML~\citep{vanschoren2014openml}:
    \begin{itemize}
        \item Suite~297 (OpenML-CTR23): a community-curated tabular regression
                benchmark~\citep{fischer2023openml};
        \item Suite~299: an additional OpenML regression collection;
        \item Suite~269: an OpenML regression collection collating further
                community-vetted datasets.
    \end{itemize}
    These suites are fetched programmatically via the OpenML Python API and
    serve as the base layer of our collection.

    \item \textbf{PMLB (Penn Machine Learning Benchmarks).}
    Regression datasets from PMLB~\citep{olson2017pmlb}, distributed as
    compressed TSV files on GitHub.
    PMLB provides a standardised, version-controlled archive of datasets
    widely used in AutoML research.

    \item \textbf{KEEL (Knowledge Extraction based on Evolutionary Learning)}
    Regression datasets from the KEEL data-mining
    repository~\citep{derrac2015keel}, distributed as zipped \texttt{.dat}
    files.
    KEEL datasets add coverage of domains less represented in OpenML and PMLB,
    such as financial time series and energy consumption.

    \item \textbf{TALENT / OpenML verified regression datasets.}
    Additional OpenML datasets drawn from the TALENT
    benchmark~\citep{liu2025talent} and independently verified as regression tasks.
    This set expands domain coverage to areas including real estate, book
    reviews, sensor fusion, and sports analytics.
\end{enumerate}

\subsection{Dataset construction and filtering}

\subsubsection{Deduplication Procedure}
\label{app:datasets:dedup}

A central challenge in multi-source dataset aggregation is that the same
underlying data frequently appears under different names, numeric identifiers,
or preprocessing variants.
To avoid having highly correlated (identical) datasets, we apply a four-level deduplication strategy:

\begin{enumerate}
    \item \textbf{Exact normalised match.}
    Dataset names are normalised by lowercasing, stripping leading numeric
    prefixes (e.g., \texttt{197\_cpu\_act} $\to$ \texttt{cpuact}), removing
    separators and special characters, and collapsing repeated characters.
    Datasets whose normalised names are identical are treated as duplicates.

    \item \textbf{Substring match.}
    For names longer than three characters, we check whether either normalised
    name is a substring of the other.
    This catches common patterns such as \texttt{housing} matching
    \texttt{californiahousing}.

    \item \textbf{Fuzzy match.}
    We compute a pairwise similarity ratio (via Python's
    \texttt{SequenceMatcher}) and flag pairs with $\geq 85\,\%$ similarity as
    duplicates.

    \item \textbf{Explicit deduplication keys.}
    Many datasets are annotated with known aliases (e.g.,
    \texttt{cpu\_act} $\leftrightarrow$ \texttt{cpuact}).
    All checks above are re-applied against these keys.
\end{enumerate}

When a duplicate is detected, we only retain one version and discard the others,
using the (arbitrary) precedence order ({OpenML $>$ PMLB $>$ KEEL $>$ scikit-learn}).

\subsubsection{Validation Filters}
\label{app:datasets:validation}

We exclude Santander\_transaction\_value (OpenML ID 42572) because of excessive runtimes for some models, including RealMLP with hyperparameter optimization.
After deduplication and this exclusion, the remaining candidate datasets are downloaded and subjected to the following checks:

\begin{itemize}
    \item \textbf{Low-cardinality target removal.}
    For successfully loaded targets, datasets with fewer than four distinct non-missing values are excluded.
    Thus, constant, binary, and three-valued targets are all removed. This is a target-cardinality screening heuristic, not a definitive classification-versus-regression test.
    In the reported benchmark, this excludes the WLAN 3D-estimation dataset (OpenML ID~45720) and \texttt{chscase\_foot} (OpenML ID~703), each with three distinct target values.
    Starting from 104 deduplicated candidates, excluding Santander and these two datasets leaves 101 datasets.

    \item \textbf{Target integrity.}
    We favour datasets derived from genuine data-collection processes. Purely synthetic or deterministic-function datasets are included only when they appear in
    established regression benchmark suites and provide useful diversity in feature--target relationships. 
    We exclude datasets which have non-tabular modalities (images, raw text, audio) and only accept datasets with pure tabular representation.
    Rows with missing or non-numeric target values are removed. If the target column is stored as a string or categorical type, we convert it to numeric and drop any rows where conversion fails.
\end{itemize}

\subsection{Preprocessing}
\label{app:datasets:preprocessing}

All datasets undergo lightweight, uniform preprocessing to ensure
compatibility with all benchmark models:
\begin{itemize}
    \item Missing numeric features are imputed with the column median.
    \item Missing categorical features are imputed with the most frequent
          category.
    \item Categorical columns are label-encoded into integer values.
    \item No feature scaling or normalisation is applied at the benchmark level;
          individual model wrappers are free to apply their own transformations
          internally.
\end{itemize}

We note that preprocessing choices---imputation strategies, categorical encoding, and any scaling---act as modelling hyperparameters that can materially affect downstream performance. 
Different model families benefit from different strategies, and several modern algorithms support native handling of missing values.
In a future release we will provide an alternative benchmark configuration that omits benchmark-level imputation, enabling direct comparison of methods that handle missing values natively.

\subsection{Final dataset overview}
\label{app:datasets:list}

Table~\ref{tab:datasets_full} enumerates the datasets in ScoringBench after
deduplication and validation.

\begin{small}
    \begin{longtable}{p{0.48\textwidth} | p{0.48\textwidth}}
    \caption{The 101 datasets included in the reported ScoringBench results.}
                               \label{tab:datasets_full} \\
                               \toprule
                               \textbf{Dataset} & \textbf{Dataset} \\
                               \midrule
                               \endfirsthead
                               \multicolumn{2}{c}{\tablename\ \thetable\ -- \textit{continued from previous page}} \\
                               \toprule
                               \textbf{Dataset} & \textbf{Dataset} \\
                               \midrule
                               \endhead
                               \midrule
                               \multicolumn{2}{r}{\textit{continued on next page}} \\
                               \endfoot
                               \bottomrule
                               \endlastfoot
                               \addlinespace
1000-Cameras-Dataset & 1027\_ESL \\
1028\_SWD & 1029\_LEV \\
1030\_ERA & 1193\_BNG\_lowbwt \\
1199\_BNG\_echoMonths & 1201\_BNG\_breastTumor \\
1203\_BNG\_pwLinear & 215\_2dplanes \\
218\_house\_8L & 225\_puma8NH \\
227\_cpu\_small & 294\_satellite\_image \\
344\_mv & 503\_wind \\
529\_pollen & 547\_no2 \\
564\_fried & 595\_fri\_c0\_1000\_10 \\
AIRFOIL & Ailerons \\
Airlines\_DepDelay\_10M & Allstate\_Claims\_Severity \\
Another-Dataset-on-used-Fiat-500-(1538-rows) & BIAS\_CORRECTION \\
BNG(mv) & BNG(stock) \\
Bike\_Sharing\_Demand & Brazilian\_houses \\
Buzzinsocialmedia\_Twitter & CPS1988 \\
CookbookReviews & Ele2 \\
Energy\_Efficiency & Goodreads-Computer-Books \\
IEEE80211aa-GATS & Infrared\_Thermography\_Temperature \\
Job\_Profitability & Laser \\
MIP-2016-regression & Mercedes\_Benz\_Greener\_Manufacturing \\
MiamiHousing2016 & Moneyball \\
Mortgage & NASA\_PHM2008 \\
OnlineNewsPopularity & PRODUCTIVITY \\
Parkinsons\_Telemonitoring & QSAR-TID-10980 \\
QSAR-TID-11 & QsarFishToxicity \\
SAT11-HAND-runtime-regression & SGEMM\_GPU\_kernel\_performance \\
Student\_Performance & Treasury \\
Wizmir & Yolanda \\
abalone & analcatdata\_supreme \\
avocado\_sales & bank32nh \\
bank8FM & black\_friday \\
boston & california \\
colleges & concrete\_compressive\_strength \\
cpu\_act & dataset\_sales \\
debutanizer & diamonds \\
elevators & fifa \\
house\_16H & house\_prices\_nominal \\
house\_sales & houses \\
isolet & kin8nm \\
mauna-loa-atmospheric-co2 & medical\_charges \\
nyc-taxi-green-dec-2016 & particulate-matter-ukair-2017 \\
pol & puma32H \\
quake & sensory \\
socmob & space\_ga \\
stock\_fardamento02 & sulfur \\
superconduct & tecator \\
topo\_2\_1 & us\_crime \\
visualizing\_soil & weather\_izmir \\
wine\_quality & year \\
yprop\_4\_1 & \\
\end{longtable}
\end{small}

\noindent

The final number of datasets used in any given experiment depends on
which candidates survive all deduplication and validation steps.
The authoritative, machine-readable dataset list is exported as
\texttt{datasets.json} at the start of every benchmark run. 
If any datasets fails on too many models (e.g., due to memory errors), 
it is automatically removed from the final results to prevent skewing the analysis.

\section{Experimental protocol}

\subsection{Cross-Validation Protocol}
\label{app:datasets:cv}

Each dataset is evaluated using $K$-fold cross-validation (CV) with
$K = 5$ folds and a fixed random seed (\texttt{seed\,=\,42}) for
reproducibility.
The number of CV repeats defaults to~1 but can be increased with
\texttt{-{}-n\_repeats\_cv} to obtain tighter confidence estimates.

\subsection{Downsampling after CV split}

To ensure computational feasibility and to prevent very large datasets from
dominating the benchmark, each cross-validation repeat draws a random
subsample of at most $N_{\max} = 3000 = 2400 + 600$ (train/test) observations (configurable via
\texttt{-{}-sample\_size}).
This design may slightly favor in-context learners, 
as more traditional parametric baselines (e.g., XGBoost, MLPs) cannot fully exploit their scaling behavior on larger datasets. \\
Subsampling is performed inside the CV loop so that each repeat sees a
different random subset, improving robustness through less correlation in the folds.
Future versions of the benchmark may study sample efficiency by measuring how model performance scales with training-set size; this requires evaluating multiple training-sample sizes per dataset.
As an ablation, we also evaluate the effect of capping the total dataset size (training plus test) at 1000, 2000 and 8000 observations, see \Cref{sec:ablation_dataset_size}.

\subsection{Evaluation}
For each CV split and each model, we compute the fold-averaged score $\bar{s}_{m,d,p}$ on held out data (from the CV split) for each metric $p$ on dataset $d$.

\subsection{Experimental runtime}
\label{subsec:experimental_runtime}

All experiments were conducted on a cluster of 40 GPU nodes equipped with NVIDIA RTX 6000 GPUs, with a total runtime of approximately 36 hours. 
A significant amount of the computational cost is attributed to \textit{RealMLP}, which is considerably slower than other methods due to its extensive hyperparameter optimization. 
Further details on training and inference times can be found in the online leaderboard.

\section{Ranking Approaches}
\label{app:details_ranking_methodology}

\subsection{Dem\v{s}ar--autorank approach (Autorank)}
\label{subsec:autorank}

Autorank strictly follows the Dem\v{s}ar protocol \citep{demsar2006statistical} and relies on the implementation in the \texttt{autorank} library \citep{Herbold2020}.
This approach discards score magnitudes and operates solely on ordinal ranks. Let $m \in \{1, \dots, M\}$ denote a model, $d \in \{1, \dots, D\}$ a dataset, and $p$ a specific performance metric. The procedure is as follows \citep{demsar2006statistical}:

\begin{itemize}
\item For each dataset $d$ and metric $p$, the models $m$ are ranked based on their fold-averaged scores $\bar{s}_{m, d, p}$. The best-performing model receives rank 1, the second-best rank 2, and so on. This transforms the data into the shape $(\text{model } m,\; \text{dataset } d,\; \text{rank } r_{m,d,p})$.\footnote{In contrast to the scoring rules, minimizing coverage is not desirable but we need to judge goodness by its agreement with the nominal level.
For ranking we, therefore, have to transform coverage and we rank according to $| \mathrm{Cov}_{1-\alpha} - (1-\alpha) |$ instead of $\mathrm{Cov}_{1-\alpha}$.
Note that the absolute value is taken per fold, before averaging over folds.}
\item Next, the mean rank $\bar{r}_{m,p}$ for each model $m$ on metric $p$ is computed by averaging across all $D$ datasets, 
yielding the mean rank 
\begin{align}
\label{eq:mean_rank}
\bar{r}_{m,p} = \frac{1}{D}\sum_{d=1}^{D} r_{m,d,p}
\end{align}
\item The non-parametric Friedman test then evaluates whether these mean ranks $\bar{r}_{m,p}$ differ significantly across the $M$ models for a given metric $p$.
\item If the omnibus Friedman test is significant at level $\alpha=0.05$, the post-hoc Nemenyi test \citep{nemenyi1963distribution} identifies which specific pairs of models differ. The Nemenyi test defines a critical difference (CD) threshold \citep{demsar2006statistical}:
\begin{align}
\label{eq:CD_value}
    \mathrm{CD} = q_\alpha \sqrt{\frac{M(M+1)}{6D}}\,,
\end{align}
where $q_\alpha$ is the critical value from the studentised range distribution divided by $\sqrt{2}$.
\item Two models are considered statistically significantly different if the absolute difference between their mean ranks exceeds the value $\mathrm{CD}$ computed in Eq.~\eqref{eq:CD_value}.
\end{itemize}

Autorank automatically selects appropriate summary statistics, confidence-interval methods, and effect-size measures for reporting. Furthermore, it automatically creates critical difference (CD) diagrams like the one in \Cref{fig:cd_diagram_crps}.

To partially mitigate the loss of magnitude information inherent in rank-based tests, \texttt{autorank} augments each model's evaluation with Akinshin's~$\gamma$, a non-parametric effect size analogous to Cohen's~$d$.
Akinshin's~$\gamma$ quantifies the standardized difference between two score distributions without parametric assumptions; we report it computed against the top-ranked model.

\subsection{Standardised magnitude-based ranking ($z$-score ranking)}
\label{subsec:app:zscore}

The second approach retains performance magnitude information by standardising the fold-averaged scores $\bar{s}_{m,d,p}$ (see Eq.~\eqref{eq:zscore_def}) within each dataset before aggregating them. 
Starting with the data in the shape $(\text{model } m,\;\text{dataset } d,\;\text{mean score } \bar{s}_{m,d,p})$, we compute a $z$-score to normalize for dataset difficulty:
\begin{equation}
\label{eq:zscore_def}
    z_{m,d,p} \;=\; \frac{\bar{s}_{m,d,p} - \mu_{d,p}}{\sigma_{d,p} + \epsilon}\,,
\end{equation}
where the dataset- and metric-specific mean $\mu_{d,p}$ and standard deviation $\sigma_{d,p}$ are defined across the $M$ models as:
\begin{align*}
    \mu_{d,p} = \frac{1}{M}\sum_{m=1}^{M} \bar{s}_{m,d,p}\,, \qquad \sigma_{d,p} = \sqrt{\frac{1}{M-1}\sum_{m=1}^{M}\bigl(\bar{s}_{m,d,p}-\mu_{d,p}\bigr)^2}\,.
\end{align*}
Here, $\epsilon=10^{-9}$ prevents division by zero, and $\sigma_{d,p}$ is estimated with $M-1$ degrees of freedom. This standardisation transforms the data into the shape:
\begin{align*}
    (\text{model } m,~\text{dataset } d,~z\text{-score } z_{m,d,p})\,,
\end{align*}
expressing each model's performance in units of the cross-model standard deviation on dataset~$d$. Crucially, this standardisation centers and scales scores within each dataset (subtracting the dataset mean and dividing by the cross-model standard deviation), removing dataset-specific location and scale so that resulting $z_{m,d,p}$ values reflect relative performance across models independent of metric scale or dataset difficulty.

For metrics $p$ where lower values indicate better performance, the sign of $z_{m,d,p}$ is negated so that higher standardised scores uniformly represent superior performance (e.g. for CRPS, lower is better, so we use $-z_{m,d,p}$ instead of $z_{m,d,p}$).
The final aggregated ranking score for model~$m$ on metric~$p$ is obtained by averaging its standardised $z$-scores across all $D$ datasets:
\begin{align*}
    \bar{z}_{m,p} \;=\; \frac{1}{D}\sum_{d=1}^{D} z_{m,d,p}\,.
\end{align*}
Models are then ranked in descending order based on $\bar{z}_{m,p}$.

We compare both ranking approaches in \Cref{fig:ranking_comparison}. 
The figure shows, for every model, how many times it was placed Top\-1 and how many times it was placed in the Top\-3 under each approach. 
Similar Top\-1 and Top\-3 counts across the two methods indicate stability ---
the same models are repeatedly identified as winners and leaders; 
large differences imply the ranking choice materially changes which models are considered top performers.
Additionally, a Pearson correlation between the two rank vectors is computed to quantify their
agreement, 
see \Cref{subsec:ranking_stability}.

\FloatBarrier
\subsection{Ranking stability between both ranking approaches}
\label{subsec:ranking_stability}

\FloatBarrier
\begin{landscape}
\begin{figure}
\centering
\includegraphics[width=1.45\textwidth]{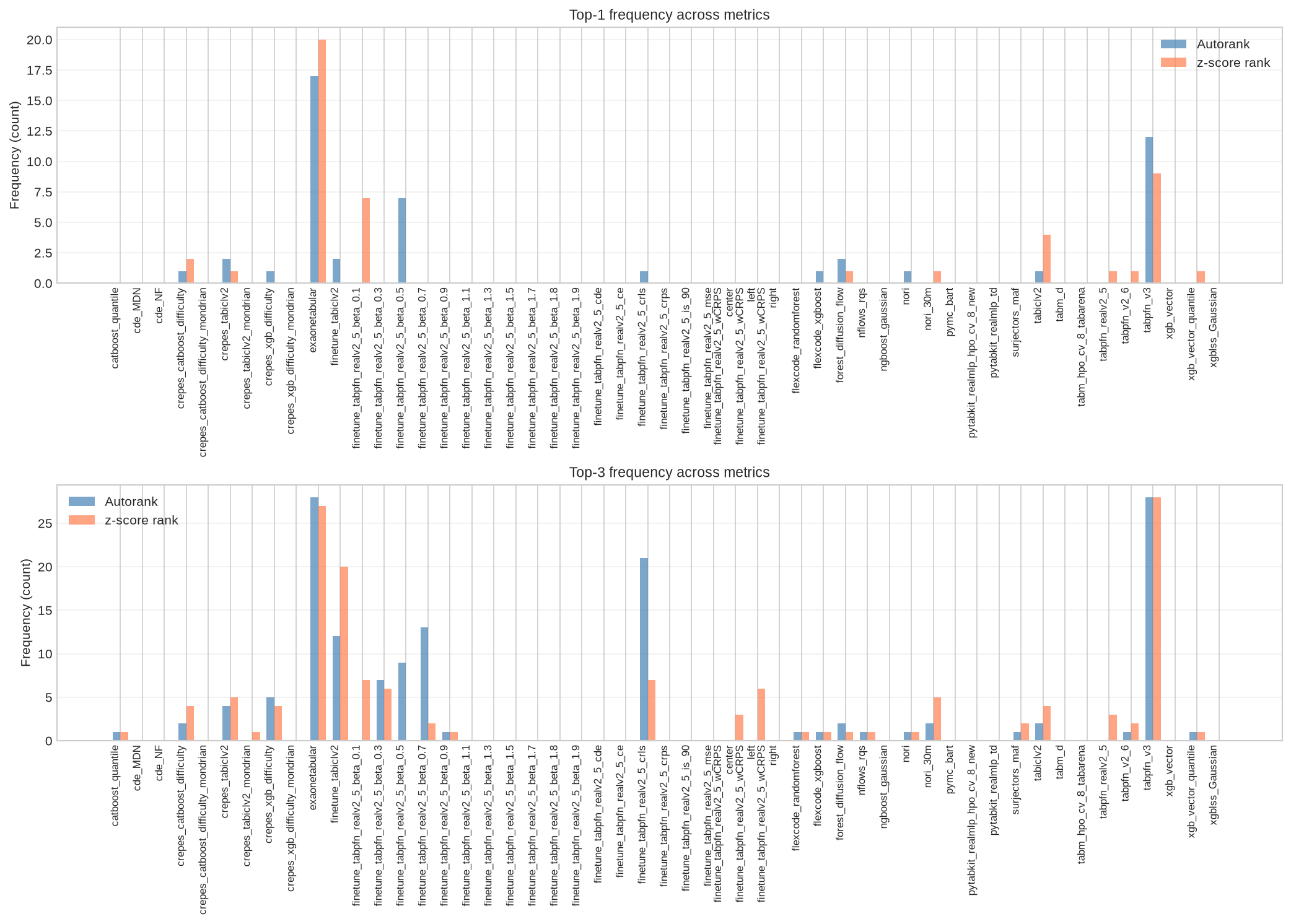}
\caption{Winner stability between ranking methods. Evaluation of Top-1 and Top-3 agreement as described in the text.}
\label{fig:ranking_comparison}
\end{figure}
\end{landscape}

To assess whether our conclusions depend on the ranking method, we compare Autorank and $z$-score rankings across all metrics. For each metric, we record 
which model is ranked first and which models appear in the top three under each method (\Cref{fig:ranking_comparison}).

The two approaches largely agree: across the $44$ metrics reported in \Cref{tab:metric_correlations} as indicated by the high Pearson correlations.

Given this overall agreement, reporting only the Autorank-based leaderboards 
in the main text is justified; both rankings for every metric are available on the live leaderboard at \url{https://scoringbench.com/}.

\input{Figures/metric_correlations_table.tex}

\FloatBarrier
\subsection{Additional ranking results}
\label{subsec:app:additional_rankings}

In the main text, the CRPS-based leaderboard was reported in \Cref{tab:leaderboard_crps}. 
Here we report the Autorank leaderboards for RMSE and the 90\% interval score
(\Cref{tab:leaderboard_rmse,tab:leaderboard_is90}).
RMSE provides the point-estimate comparison to CRPS; the interval score adds an interval-based view of distributional quality.
The $z$-score-based leaderboards are available at \url{https://scoringbench.com/}.
These tables match the structure of \Cref{tab:leaderboard_crps}.

\begin{leaderboardtable}
{Autorank leaderboard for CRPS. Columns: MR = mean rank across datasets (lower is better); MED = median score; MAD = median absolute deviation; CI = confidence interval for the median; $\gamma$ = Akinshin's gamma effect size relative to the top-ranked model. \textit{Note:} MED, MAD, and $\gamma$ are computed by autorank without further standardization, thus reflecting the original score scale.
While this may hamper direct comparison across datasets with different units, we report it for transparency; ranking is unaffected as autorank sorts by mean rank (scale-independent). The $z$-score ranking approach properly standardizes for dataset difficulty and produces similar ordering as the autorank approach due to high correlation between both methods (Appendix~\ref{subsec:ranking_stability}) while having interpretable z-scores; Live rankings available at \url{https://scoringbench.com/}.}
{tab:leaderboard_crps}
\input{Figures/output_3000/figures/leaderboard/latex_table_crps.tex}
\end{leaderboardtable}

\begin{leaderboardtable}
{Autorank leaderboard for the 90\% interval score (lower is better). Column definitions as in \Cref{tab:leaderboard_crps}.}
{tab:leaderboard_is90}
\input{Figures/output_3000/figures/leaderboard/latex_table_interval_score_90.tex}
\end{leaderboardtable}

\begin{leaderboardtable}
{Autorank leaderboard for RMSE (lower is better). Column definitions as in \Cref{tab:leaderboard_crps}.}
{tab:leaderboard_rmse}
\input{Figures/output_3000/figures/leaderboard/latex_table_rmse.tex}
\end{leaderboardtable}

 \FloatBarrier

\section{Ablation study with respect to dataset size} 
\label{sec:ablation_dataset_size}
\begin{landscape}
\begin{figure}
\centering
\includegraphics[width=\linewidth]{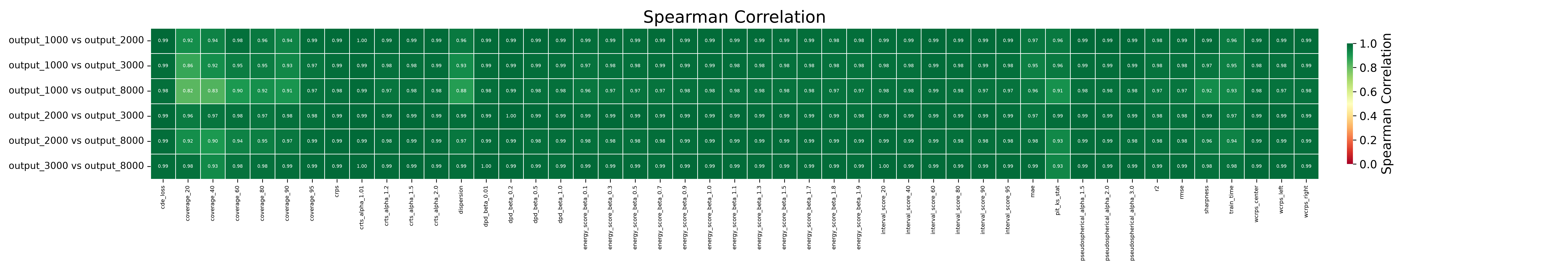}
\caption{Ablation study on dataset size.
We compare model rankings across all six pairs of sample-size caps (1000, 2000, 3000 and 8000 training plus test observations) using Spearman correlation over the 51 models common to all four runs.
The heatmap displays 47 metrics, excluding the PIT KS test $p$-value, and shows high correlations across sample sizes.}
\label{fig:ablation_dataset_size}
\end{figure}
\end{landscape}

To investigate if our ranking results are robust to the choice of dataset size, we perform an ablation study where we re-run the benchmark on all datasets but with capped dataset sizes of 1000, 2000, 3000 and 8000 samples (test + train size).
The results are shown in \Cref{fig:ablation_dataset_size}.

\FloatBarrier

We observe some minor variation across metrics and models; this is expected because different scoring rules can have different finite-sample efficiencies and may
converge at different speeds during finetuning, which can affect rankings at smaller sample sizes \citep{waghmare2025proper}.

Let $\mathcal{M}_{a,b}^{(p)}$ be the set of models common to sample-size outputs (folders) $a$ and $b$ for metric $p$, where $n_{a,b}^{(p)} = |\mathcal{M}_{a,b}^{(p)}|$. For each model $i \in \mathcal{M}_{a,b}^{(p)}$, let $r^{(a,p)}_i$ and $r^{(b,p)}_i$ be its ranks in folders $a$ and $b$ for metric $p$. The Spearman rank correlation $\rho^{(p)}_{a,b}$ is defined as:

\[
\rho^{(p)}_{a,b} = \frac{\sum_{i=1}^{n_{a,b}^{(p)}} \left(r^{(a,p)}_i - \bar{r}^{(a,p)}\right) \left(r^{(b,p)}_i - \bar{r}^{(b,p)}\right)}{\sqrt{\sum_{i=1}^{n_{a,b}^{(p)}} \left(r^{(a,p)}_i - \bar{r}^{(a,p)}\right)^2} \sqrt{\sum_{i=1}^{n_{a,b}^{(p)}} \left(r^{(b,p)}_i - \bar{r}^{(b,p)}\right)^2}}
\]

where $\bar{r}^{(a,p)}$ and $\bar{r}^{(b,p)}$ are the mean ranks. Ties are handled by assigning average ranks.

\FloatBarrier




\section{Workflow for contributing to the benchmark}

In order to facilitate contributions to our benchmark from other researchers,
we briefly elaborate on the workflow to do so.

\begin{enumerate}
\item \texttt{git clone --recurse-submodules} \url{https://github.com/jonaslandsgesell/ScoringBench.git}, make sure to have git-lfs installed.
\item Add your custom wrapper with a unique name to the folder \texttt{scoringbench/wrapper/} following the template of existing wrappers. Ensure it adheres to the expected interface for seamless integration.
\item Run the benchmarking via python \texttt{run\_bench\_regression.py} (or parallelize via slurm with \texttt{sbatch \-\-array=0\-103 run\_benchmark.sbatch})
\item Run the ranking method by executing \texttt{python autorank\_leaderboard.py} which will evaluate both Autorank and $z$-score ranking and generate the leaderboard outputs.
\item Commit your model parquet file (documenting each run) and the updated leaderboard CSVs in \texttt{/output/figures/leaderboard/}. Since the output repository is separate from the main repository, push to both. This serves as a public ledger and allows traceability.
\item Create a pull request to the ScoringBench repository for review; contributions that meet standards will be merged.
\item Upon merge, \url{https://scoringbench.com/} will automatically display the updated leaderboard and data is archived in the git lfs repository for reproducibility.
\end{enumerate}

\section{Broader Impact}
\label{sec:broader_impact}

Our work evaluates tabular foundation models, their fine-tuned variants, and additional baseline methods under a range of proper scoring rules, 
and compiles the results into a comprehensive benchmark. 
While our findings highlight that the choice of evaluation metric constitutes an important modeling decision with potential implications across various applications, 
we do not identify any specific broader impacts that require particular emphasis.

%% file: Figures/metric_correlations_table.tex
\begin{table}[ht]
\centering
\caption{Pearson correlation of ranks between Autorank and Zscore ranking across metrics (computed over $51$ models).}
\label{tab:metric_correlations}
\begin{tabular}{lc}
\hline
\textbf{Metric} & \textbf{Rank Correlation ($r$)} \\
\hline
cde\_loss & 0.9743 \\
crps & 0.9910 \\
dispersion & 0.9680 \\
dpd\_beta\_0.01 & 0.9880 \\
dpd\_beta\_0.2 & 0.9720 \\
dpd\_beta\_0.5 & 0.9716 \\
dpd\_beta\_1.0 & 0.9743 \\
energy\_score\_beta\_0.1 & 0.9820 \\
energy\_score\_beta\_0.3 & 0.9871 \\
energy\_score\_beta\_0.5 & 0.9924 \\
energy\_score\_beta\_0.7 & 0.9925 \\
energy\_score\_beta\_0.9 & 0.9920 \\
energy\_score\_beta\_1.0 & 0.9910 \\
energy\_score\_beta\_1.1 & 0.9902 \\
energy\_score\_beta\_1.3 & 0.9890 \\
energy\_score\_beta\_1.5 & 0.9664 \\
energy\_score\_beta\_1.7 & 0.9507 \\
energy\_score\_beta\_1.8 & 0.9500 \\
energy\_score\_beta\_1.9 & 0.9460 \\
interval\_score\_20 & 0.9878 \\
interval\_score\_40 & 0.9879 \\
interval\_score\_60 & 0.9879 \\
interval\_score\_80 & 0.9894 \\
interval\_score\_90 & 0.9825 \\
interval\_score\_95 & 0.9768 \\
mae & 0.9716 \\
pit\_ks\_pvalue & 0.9298 \\
pit\_ks\_stat & 0.8119 \\
pseudospherical\_alpha\_1.5 & 0.9757 \\
pseudospherical\_alpha\_2.0 & 0.9773 \\
pseudospherical\_alpha\_3.0 & 0.9836 \\
r2 & 0.9593 \\
rmse & 0.9539 \\
sharpness & 0.9636 \\
train\_time & 0.9720 \\
wcrps\_center & 0.9919 \\
wcrps\_left & 0.9892 \\
wcrps\_right & 0.9906 \\
\hline
\end{tabular}
\end{table}

%% file: Figures/output_3000/figures/leaderboard/latex_table_crps.tex
\begin{tabular}{lrrllll}
\toprule
 & MR & MED & MAD & CI & $\gamma$ & Magnitude \\
\midrule
exaonetabular & 8.723 & 0.378 & 0.362 & [0.073, 1.597] & 0.000 & negligible \\
tabpfn\_v3 & 8.733 & 0.373 & 0.356 & [0.092, 1.608] & 0.010 & negligible \\
finetune\_tabpfn\_realv2\_5\_crls & 11.084 & 0.377 & 0.363 & [0.091, 1.652] & 0.001 & negligible \\
finetune\_tabiclv2 & 11.871 & 0.392 & 0.377 & [0.092, 1.600] & -0.025 & negligible \\
finetune\_tabpfn\_realv2\_5\_beta\_0.7 & 12.386 & 0.379 & 0.364 & [0.092, 1.651] & -0.002 & negligible \\
finetune\_tabpfn\_realv2\_5\_beta\_0.5 & 12.569 & 0.379 & 0.365 & [0.092, 1.651] & -0.003 & negligible \\
finetune\_tabpfn\_realv2\_5\_crps & 12.698 & 0.381 & 0.367 & [0.092, 1.649] & -0.006 & negligible \\
finetune\_tabpfn\_realv2\_5\_beta\_1.1 & 12.871 & 0.381 & 0.367 & [0.092, 1.651] & -0.007 & negligible \\
finetune\_tabpfn\_realv2\_5\_beta\_0.9 & 12.911 & 0.381 & 0.366 & [0.092, 1.650] & -0.006 & negligible \\
finetune\_tabpfn\_realv2\_5\_beta\_0.3 & 13.663 & 0.379 & 0.365 & [0.092, 1.651] & -0.003 & negligible \\
finetune\_tabpfn\_realv2\_5\_beta\_1.3 & 14.015 & 0.382 & 0.367 & [0.092, 1.656] & -0.007 & negligible \\
tabiclv2 & 14.119 & 0.392 & 0.378 & [0.096, 1.602] & -0.026 & negligible \\
nori\_30m & 14.337 & 0.393 & 0.377 & [0.074, 1.621] & -0.028 & negligible \\
finetune\_tabpfn\_realv2\_5\_mse & 14.817 & 0.378 & 0.363 & [0.092, 1.653] & -0.000 & negligible \\
finetune\_tabpfn\_realv2\_5\_beta\_1.5 & 14.980 & 0.382 & 0.368 & [0.092, 1.656] & -0.008 & negligible \\
finetune\_tabpfn\_realv2\_5\_beta\_0.1 & 15.030 & 0.379 & 0.365 & [0.092, 1.652] & -0.003 & negligible \\
finetune\_tabpfn\_realv2\_5\_wCRPS\_center & 15.069 & 0.380 & 0.366 & [0.092, 1.654] & -0.005 & negligible \\
finetune\_tabpfn\_realv2\_5\_wCRPS\_right & 15.851 & 0.384 & 0.369 & [0.092, 1.647] & -0.012 & negligible \\
finetune\_tabpfn\_realv2\_5\_beta\_1.7 & 16.139 & 0.382 & 0.367 & [0.092, 1.655] & -0.008 & negligible \\
finetune\_tabpfn\_realv2\_5\_beta\_1.8 & 16.416 & 0.382 & 0.368 & [0.092, 1.656] & -0.008 & negligible \\
finetune\_tabpfn\_realv2\_5\_wCRPS\_left & 16.510 & 0.387 & 0.373 & [0.092, 1.653] & -0.017 & negligible \\
finetune\_tabpfn\_realv2\_5\_beta\_1.9 & 16.797 & 0.383 & 0.369 & [0.092, 1.655] & -0.010 & negligible \\
finetune\_tabpfn\_realv2\_5\_cde & 17.589 & 0.381 & 0.366 & [0.092, 1.653] & -0.005 & negligible \\
tabpfn\_realv2\_5 & 18.208 & 0.378 & 0.363 & [0.080, 1.655] & -0.001 & negligible \\
finetune\_tabpfn\_realv2\_5\_ce & 18.441 & 0.381 & 0.366 & [0.092, 1.655] & -0.005 & negligible \\
finetune\_tabpfn\_realv2\_5\_is\_90 & 19.728 & 0.381 & 0.366 & [0.092, 1.658] & -0.006 & negligible \\
nori & 20.941 & 0.396 & 0.379 & [0.074, 1.639] & -0.034 & negligible \\
tabpfn\_v2\_6 & 22.911 & 0.394 & 0.378 & [0.105, 1.652] & -0.030 & negligible \\
pytabkit\_realmlp\_hpo\_cv\_8\_new & 25.693 & 0.401 & 0.386 & [0.096, 1.658] & -0.042 & negligible \\
tabm\_hpo\_cv\_8\_tabarena & 28.000 & 0.426 & 0.397 & [0.097, 1.679] & -0.086 & negligible \\
crepes\_tabiclv2 & 31.960 & 0.408 & 0.391 & [0.099, 1.730] & -0.054 & negligible \\
crepes\_tabiclv2\_mondrian & 33.723 & 0.429 & 0.412 & [0.100, 1.752] & -0.090 & negligible \\
pytabkit\_realmlp\_td & 34.485 & 0.451 & 0.423 & [0.108, 1.787] & -0.125 & negligible \\
tabm\_d & 34.851 & 0.442 & 0.427 & [0.122, 2.076] & -0.110 & negligible \\
catboost\_quantile & 36.594 & 0.482 & 0.462 & [0.106, 2.381] & -0.169 & negligible \\
xgb\_vector & 39.030 & 0.475 & 0.449 & [0.131, 2.385] & -0.160 & negligible \\
flexcode\_randomforest & 39.614 & 0.482 & 0.461 & [0.120, 2.354] & -0.169 & negligible \\
ngboost\_gaussian & 39.782 & 0.473 & 0.451 & [0.128, 3.046] & -0.157 & negligible \\
crepes\_catboost\_difficulty & 40.851 & 0.472 & 0.450 & [0.126, 2.591] & -0.155 & negligible \\
cde\_MDN & 41.317 & 0.560 & 0.524 & [0.125, 2.353] & -0.273 & small \\
forest\_diffusion\_flow & 41.436 & 0.475 & 0.449 & [0.119, 2.429] & -0.161 & negligible \\
xgb\_vector\_quantile & 41.980 & 0.514 & 0.491 & [0.119, 2.606] & -0.213 & small \\
xgblss\_Gaussian & 42.119 & 0.540 & 0.510 & [0.131, 3.165] & -0.247 & small \\
flexcode\_xgboost & 42.327 & 0.563 & 0.527 & [0.141, 2.578] & -0.276 & small \\
cde\_NF & 42.851 & 0.606 & 0.556 & [0.138, 2.308] & -0.328 & small \\
crepes\_xgb\_difficulty & 42.901 & 0.467 & 0.446 & [0.118, 2.683] & -0.148 & negligible \\
crepes\_catboost\_difficulty\_mondrian & 43.683 & 0.496 & 0.477 & [0.136, 2.768] & -0.188 & negligible \\
nflows\_rqs & 44.941 & 0.548 & 0.513 & [0.155, 2.411] & -0.259 & small \\
crepes\_xgb\_difficulty\_mondrian & 45.584 & 0.506 & 0.484 & [0.122, 2.810] & -0.202 & small \\
surjectors\_maf & 46.079 & 0.564 & 0.528 & [0.152, 2.577] & -0.278 & small \\
pymc\_bart & 46.792 & 0.894 & 0.837 & [0.190, 4.376] & -0.540 & medium \\
\bottomrule
\end{tabular}

%% file: Figures/output_3000/figures/leaderboard/latex_table_interval_score_90.tex
\begin{tabular}{lrrllll}
\toprule
 & MR & MED & MAD & CI & $\gamma$ & Magnitude \\
\midrule
exaonetabular & 8.604 & 2.963 & 2.816 & [0.663, 12.150] & 0.000 & negligible \\
tabpfn\_v3 & 10.327 & 3.007 & 2.811 & [0.690, 12.049] & -0.011 & negligible \\
finetune\_tabpfn\_realv2\_5\_crls & 10.678 & 2.968 & 2.833 & [0.688, 12.285] & -0.001 & negligible \\
finetune\_tabiclv2 & 11.020 & 3.006 & 2.879 & [0.689, 12.238] & -0.010 & negligible \\
tabiclv2 & 12.584 & 2.997 & 2.852 & [0.685, 12.168] & -0.008 & negligible \\
finetune\_tabpfn\_realv2\_5\_beta\_0.7 & 13.485 & 3.001 & 2.824 & [0.692, 12.252] & -0.009 & negligible \\
nori\_30m & 13.535 & 2.988 & 2.832 & [0.601, 12.407] & -0.006 & negligible \\
finetune\_tabpfn\_realv2\_5\_beta\_0.5 & 13.569 & 3.001 & 2.839 & [0.691, 12.258] & -0.009 & negligible \\
finetune\_tabpfn\_realv2\_5\_beta\_0.3 & 13.614 & 3.002 & 2.840 & [0.691, 12.259] & -0.009 & negligible \\
finetune\_tabpfn\_realv2\_5\_crps & 13.688 & 3.002 & 2.837 & [0.692, 12.261] & -0.009 & negligible \\
finetune\_tabpfn\_realv2\_5\_beta\_0.9 & 14.198 & 2.998 & 2.833 & [0.692, 12.255] & -0.008 & negligible \\
finetune\_tabpfn\_realv2\_5\_beta\_1.1 & 14.277 & 3.004 & 2.838 & [0.692, 12.259] & -0.010 & negligible \\
finetune\_tabpfn\_realv2\_5\_wCRPS\_right & 14.465 & 3.028 & 2.864 & [0.692, 12.291] & -0.016 & negligible \\
finetune\_tabpfn\_realv2\_5\_beta\_1.3 & 14.470 & 3.017 & 2.851 & [0.692, 12.269] & -0.013 & negligible \\
finetune\_tabpfn\_realv2\_5\_beta\_1.5 & 14.505 & 3.029 & 2.864 & [0.691, 12.276] & -0.016 & negligible \\
finetune\_tabpfn\_realv2\_5\_beta\_0.1 & 14.851 & 3.012 & 2.850 & [0.691, 12.270] & -0.012 & negligible \\
finetune\_tabpfn\_realv2\_5\_is\_90 & 15.530 & 3.020 & 2.858 & [0.691, 12.289] & -0.014 & negligible \\
finetune\_tabpfn\_realv2\_5\_mse & 15.876 & 3.063 & 2.876 & [0.692, 12.330] & -0.024 & negligible \\
finetune\_tabpfn\_realv2\_5\_beta\_1.7 & 16.198 & 3.067 & 2.903 & [0.691, 12.285] & -0.025 & negligible \\
finetune\_tabpfn\_realv2\_5\_wCRPS\_center & 16.228 & 3.043 & 2.874 & [0.691, 12.277] & -0.019 & negligible \\
finetune\_tabpfn\_realv2\_5\_wCRPS\_left & 16.431 & 3.057 & 2.890 & [0.690, 12.335] & -0.022 & negligible \\
finetune\_tabpfn\_realv2\_5\_beta\_1.8 & 16.861 & 3.057 & 2.894 & [0.690, 12.293] & -0.022 & negligible \\
finetune\_tabpfn\_realv2\_5\_beta\_1.9 & 17.629 & 3.066 & 2.904 & [0.689, 12.293] & -0.024 & negligible \\
tabpfn\_realv2\_5 & 17.802 & 3.059 & 2.877 & [0.695, 12.309] & -0.023 & negligible \\
finetune\_tabpfn\_realv2\_5\_cde & 18.114 & 3.069 & 2.907 & [0.696, 12.247] & -0.025 & negligible \\
finetune\_tabpfn\_realv2\_5\_ce & 18.391 & 3.071 & 2.909 & [0.696, 12.330] & -0.025 & negligible \\
nori & 19.228 & 3.011 & 2.848 & [0.693, 12.571] & -0.011 & negligible \\
tabpfn\_v2\_6 & 22.752 & 3.067 & 2.845 & [0.885, 13.411] & -0.025 & negligible \\
pytabkit\_realmlp\_hpo\_cv\_8\_new & 26.802 & 3.124 & 2.987 & [0.741, 15.931] & -0.038 & negligible \\
tabm\_hpo\_cv\_8\_tabarena & 27.980 & 3.528 & 3.263 & [0.781, 18.557] & -0.125 & negligible \\
pytabkit\_realmlp\_td & 33.069 & 3.678 & 3.499 & [0.944, 17.701] & -0.152 & negligible \\
crepes\_tabiclv2 & 33.743 & 3.764 & 3.522 & [0.797, 19.249] & -0.169 & negligible \\
tabm\_d & 34.119 & 3.609 & 3.411 & [1.012, 20.939] & -0.139 & negligible \\
crepes\_tabiclv2\_mondrian & 35.772 & 3.876 & 3.695 & [0.866, 20.670] & -0.188 & negligible \\
ngboost\_gaussian & 36.723 & 3.816 & 3.629 & [1.161, 24.067] & -0.177 & negligible \\
xgb\_vector & 36.851 & 4.619 & 4.334 & [1.155, 20.140] & -0.306 & small \\
cde\_MDN & 37.693 & 4.640 & 4.357 & [1.058, 23.694] & -0.308 & small \\
crepes\_catboost\_difficulty & 39.020 & 4.045 & 3.834 & [1.111, 24.556] & -0.217 & small \\
cde\_NF & 40.822 & 6.031 & 5.714 & [1.471, 26.977] & -0.459 & small \\
catboost\_quantile & 40.901 & 4.686 & 4.474 & [1.087, 24.938] & -0.311 & small \\
xgblss\_Gaussian & 40.950 & 4.388 & 4.205 & [1.196, 25.127] & -0.269 & small \\
crepes\_xgb\_difficulty & 41.158 & 4.128 & 3.925 & [1.076, 25.667] & -0.230 & small \\
crepes\_catboost\_difficulty\_mondrian & 42.861 & 4.936 & 4.631 & [1.199, 39.635] & -0.347 & small \\
xgb\_vector\_quantile & 43.347 & 5.580 & 5.344 & [1.310, 30.291] & -0.413 & small \\
flexcode\_randomforest & 43.604 & 5.263 & 5.030 & [1.638, 28.275] & -0.381 & small \\
pymc\_bart & 44.030 & 7.531 & 7.125 & [1.474, 42.480] & -0.569 & medium \\
nflows\_rqs & 44.327 & 6.592 & 6.125 & [1.538, 29.673] & -0.514 & medium \\
crepes\_xgb\_difficulty\_mondrian & 44.634 & 5.257 & 4.913 & [1.288, 34.131] & -0.386 & small \\
flexcode\_xgboost & 45.604 & 6.912 & 6.715 & [2.081, 38.746] & -0.517 & medium \\
forest\_diffusion\_flow & 45.752 & 5.723 & 5.421 & [1.393, 31.649] & -0.431 & small \\
surjectors\_maf & 47.327 & 7.705 & 7.323 & [1.677, 35.535] & -0.577 & medium \\
\bottomrule
\end{tabular}

%% file: Figures/output_3000/figures/leaderboard/latex_table_rmse.tex
\begin{tabular}{lrrllll}
\toprule
 & MR & MED & MAD & CI & $\gamma$ & Magnitude \\
\midrule
exaonetabular & 8.564 & 0.758 & 0.726 & [0.181, 3.681] & 0.000 & negligible \\
tabpfn\_v3 & 9.851 & 0.763 & 0.732 & [0.183, 3.809] & -0.005 & negligible \\
finetune\_tabpfn\_realv2\_5\_crls & 12.649 & 0.774 & 0.742 & [0.180, 3.832] & -0.015 & negligible \\
finetune\_tabiclv2 & 13.287 & 0.837 & 0.805 & [0.182, 3.768] & -0.070 & negligible \\
finetune\_tabpfn\_realv2\_5\_mse & 14.104 & 0.770 & 0.738 & [0.181, 3.791] & -0.012 & negligible \\
finetune\_tabpfn\_realv2\_5\_beta\_0.5 & 14.282 & 0.770 & 0.738 & [0.181, 3.829] & -0.012 & negligible \\
finetune\_tabpfn\_realv2\_5\_beta\_0.7 & 14.644 & 0.770 & 0.738 & [0.181, 3.835] & -0.011 & negligible \\
nori\_30m & 14.881 & 0.762 & 0.727 & [0.180, 3.734] & -0.004 & negligible \\
finetune\_tabpfn\_realv2\_5\_beta\_0.3 & 15.267 & 0.773 & 0.742 & [0.181, 3.829] & -0.014 & negligible \\
finetune\_tabpfn\_realv2\_5\_beta\_0.9 & 15.683 & 0.770 & 0.738 & [0.181, 3.834] & -0.011 & negligible \\
finetune\_tabpfn\_realv2\_5\_crps & 15.807 & 0.772 & 0.741 & [0.181, 3.807] & -0.013 & negligible \\
finetune\_tabpfn\_realv2\_5\_beta\_1.1 & 15.871 & 0.773 & 0.742 & [0.181, 3.806] & -0.014 & negligible \\
tabiclv2 & 15.901 & 0.834 & 0.798 & [0.188, 4.023] & -0.068 & negligible \\
finetune\_tabpfn\_realv2\_5\_beta\_0.1 & 16.248 & 0.774 & 0.743 & [0.181, 3.829] & -0.015 & negligible \\
finetune\_tabpfn\_realv2\_5\_beta\_1.8 & 16.644 & 0.770 & 0.739 & [0.181, 3.848] & -0.011 & negligible \\
finetune\_tabpfn\_realv2\_5\_beta\_1.5 & 16.693 & 0.772 & 0.741 & [0.181, 3.812] & -0.013 & negligible \\
finetune\_tabpfn\_realv2\_5\_beta\_1.3 & 16.827 & 0.769 & 0.738 & [0.181, 3.811] & -0.011 & negligible \\
finetune\_tabpfn\_realv2\_5\_beta\_1.9 & 16.847 & 0.772 & 0.741 & [0.181, 3.846] & -0.013 & negligible \\
finetune\_tabpfn\_realv2\_5\_wCRPS\_center & 16.891 & 0.768 & 0.737 & [0.181, 3.822] & -0.009 & negligible \\
finetune\_tabpfn\_realv2\_5\_beta\_1.7 & 16.980 & 0.770 & 0.739 & [0.181, 3.839] & -0.011 & negligible \\
finetune\_tabpfn\_realv2\_5\_wCRPS\_left & 17.797 & 0.774 & 0.743 & [0.181, 3.826] & -0.015 & negligible \\
tabpfn\_v2\_6 & 18.079 & 0.772 & 0.737 & [0.182, 4.018] & -0.014 & negligible \\
finetune\_tabpfn\_realv2\_5\_wCRPS\_right & 18.436 & 0.777 & 0.746 & [0.181, 3.819] & -0.018 & negligible \\
finetune\_tabpfn\_realv2\_5\_cde & 18.520 & 0.776 & 0.745 & [0.181, 3.829] & -0.017 & negligible \\
tabpfn\_realv2\_5 & 19.267 & 0.789 & 0.752 & [0.181, 3.822] & -0.029 & negligible \\
nori & 19.614 & 0.766 & 0.732 & [0.181, 3.759] & -0.008 & negligible \\
finetune\_tabpfn\_realv2\_5\_ce & 19.619 & 0.776 & 0.745 & [0.181, 3.829] & -0.017 & negligible \\
finetune\_tabpfn\_realv2\_5\_is\_90 & 20.233 & 0.788 & 0.757 & [0.181, 3.859] & -0.027 & negligible \\
pytabkit\_realmlp\_hpo\_cv\_8\_new & 22.723 & 0.769 & 0.735 & [0.187, 3.728] & -0.011 & negligible \\
crepes\_tabiclv2 & 23.208 & 0.846 & 0.804 & [0.188, 3.886] & -0.078 & negligible \\
tabm\_hpo\_cv\_8\_tabarena & 26.713 & 0.846 & 0.773 & [0.188, 3.906] & -0.080 & negligible \\
crepes\_tabiclv2\_mondrian & 29.307 & 0.851 & 0.808 & [0.191, 3.918] & -0.082 & negligible \\
catboost\_quantile & 33.703 & 0.849 & 0.807 & [0.210, 4.982] & -0.080 & negligible \\
forest\_diffusion\_flow & 35.525 & 0.862 & 0.784 & [0.210, 5.007] & -0.093 & negligible \\
tabm\_d & 35.634 & 0.847 & 0.807 & [0.197, 4.997] & -0.079 & negligible \\
pytabkit\_realmlp\_td & 35.960 & 0.858 & 0.802 & [0.206, 4.896] & -0.089 & negligible \\
ngboost\_gaussian & 37.941 & 0.911 & 0.834 & [0.238, 5.659] & -0.132 & negligible \\
xgb\_vector & 38.970 & 0.888 & 0.823 & [0.246, 5.254] & -0.113 & negligible \\
crepes\_catboost\_difficulty & 39.198 & 0.891 & 0.839 & [0.243, 5.200] & -0.115 & negligible \\
xgb\_vector\_quantile & 39.921 & 0.898 & 0.817 & [0.229, 5.225] & -0.122 & negligible \\
xgblss\_Gaussian & 40.277 & 1.089 & 1.028 & [0.246, 6.058] & -0.251 & small \\
cde\_MDN & 42.158 & 0.969 & 0.934 & [0.250, 6.554] & -0.171 & negligible \\
crepes\_catboost\_difficulty\_mondrian & 42.218 & 0.906 & 0.862 & [0.252, 6.248] & -0.125 & negligible \\
flexcode\_randomforest & 42.406 & 0.923 & 0.851 & [0.269, 5.555] & -0.141 & negligible \\
crepes\_xgb\_difficulty & 42.752 & 0.919 & 0.866 & [0.242, 5.388] & -0.136 & negligible \\
cde\_NF & 42.772 & 0.966 & 0.928 & [0.251, 5.416] & -0.168 & negligible \\
pymc\_bart & 44.396 & 1.386 & 1.301 & [0.287, 8.595] & -0.403 & small \\
nflows\_rqs & 44.465 & 0.968 & 0.906 & [0.267, 5.539] & -0.172 & negligible \\
crepes\_xgb\_difficulty\_mondrian & 45.040 & 0.939 & 0.886 & [0.251, 5.568] & -0.151 & negligible \\
flexcode\_xgboost & 45.356 & 1.118 & 1.081 & [0.320, 6.612] & -0.264 & small \\
surjectors\_maf & 45.871 & 0.922 & 0.837 & [0.336, 5.392] & -0.142 & negligible \\
\bottomrule
\end{tabular}

%% file: references.bib
@article{dawid1984present,
  title={Present position and potential developments: Some personal views statistical theory the prequential approach},
  author={Dawid, A Philip},
  journal={Journal of the Royal Statistical Society: Series A (General)},
  volume={147},
  number={2},
  pages={278--290},
  year={1984},
  publisher={Wiley Online Library}
}

@article{diebold1998,
  author  = {Diebold, Francis X. and Gunther, Todd A. and Tay, Anthony S.},
  title   = {Evaluating Density Forecasts with Applications to Financial Risk Management},
  journal = {International Economic Review},
  volume  = {39},
  number  = {4},
  pages   = {863--883},
  year    = {1998},
  publisher = {Wiley}
}

@article{czado2009predictive,
  title={Predictive model assessment for count data},
  author={Czado, Claudia and Gneiting, Tilmann and Held, Leonhard},
  journal={Biometrics},
  volume={65},
  number={4},
  pages={1254--1261},
  year={2009},
  publisher={Wiley Online Library}
}

@article{martin2022optimal,
  title={Optimal probabilistic forecasts: When do they work?},
  author={Martin, Gael M.\ and Loaiza-Maya, Rub{\'e}n and Maneesoonthorn, Worapree and Frazier, David T.\ and Ram{\'\i}rez-Hassan, Andr{\'e}s},
  journal={International Journal of Forecasting},
  volume={38},
  number={1},
  pages={384--406},
  year={2022},
  publisher={Elsevier}
}

@article{izbicki2016nonparametric,
  title={Nonparametric conditional density estimation in a high-dimensional regression setting},
  author={Izbicki, Rafael and Lee, Ann B},
  journal={Journal of Computational and Graphical Statistics},
  volume={25},
  number={4},
  pages={1297--1316},
  year={2016},
  publisher={Taylor \& Francis}
}

@article{garg2025real,
  title={Real-{TabPFN}: Improving tabular foundation models via continued pre-training with real-world data},
  author={Garg, Anurag and Ali, Muhammad and Hollmann, Noah and Purucker, Lennart and M{\"u}ller, Samuel and Hutter, Frank},
  journal={arXiv preprint arXiv:2507.03971},
  year={2025}
}

@article{grinsztajn2025tabpfn,
  title={{TabPFN}-2.5: Advancing the state of the art in tabular foundation models},
  author={Grinsztajn, L{\'e}o and Fl{\"o}ge, Klemens and Key, Oscar and Birkel, Felix and Jund, Philipp and Roof, Brendan and J{\"a}ger, Benjamin and Safaric, Dominik and Alessi, Simone and Hayler, Adrian and others},
  journal={arXiv preprint arXiv:2511.08667},
  year={2025}
}

@article{marz2019xgboostlss,
  title={{XGBoostLSS}--An extension of {XGBoost} to probabilistic forecasting},
  author={M{\"a}rz, Alexander},
  journal={arXiv preprint arXiv:1907.03178},
  year={2019}
}

@article{prokhorenkova2018catboost,
  title={{CatBoost}: unbiased boosting with categorical features},
  author={Prokhorenkova, Liudmila and Gusev, Gleb and Vorobev, Aleksandr and Dorogush, Anna Veronika and Gulin, Andrey},
  journal={Advances in neural information processing systems},
  volume={31},
  year={2018}
}

@article{holzmuller2024better,
  title={{B}etter by {D}efault: {S}trong {P}re-{T}uned {MLP}s and {B}oosted {T}rees on {T}abular {D}ata},
  author={Holzm{\"u}ller, David and Grinsztajn, L{\'e}o and Steinwart, Ingo},
  journal={Advances in Neural Information Processing Systems},
  volume={37},
  pages={26577--26658},
  year={2024}
}

@article{gorishniy2024tabm,
  title={{TabM}: Advancing tabular deep learning with parameter-efficient ensembling},
  author={Gorishniy, Yury and Kotelnikov, Akim and Babenko, Artem},
  journal={arXiv preprint arXiv:2410.24210},
  year={2024}
}

@article{landsgesell2026distributional,
  title={Distributional regression with tabular foundation models: Evaluating probabilistic predictions via proper scoring rules},
  author={Landsgesell, Jonas and Knoll, Pascal and Wenzel, Tizian},
  journal={arXiv preprint arXiv:2603.08206},
  year={2026}
}

@article{Herbold2020,
  doi = {10.21105/joss.02173},
  url = {https://doi.org/10.21105/joss.02173},
  year = {2020},
  publisher = {The Open Journal},
  volume = {5},
  number = {48},
  pages = {2173},
  author = {Steffen Herbold},
  title = {Autorank: A Python package for automated ranking of classifiers},
  journal = {Journal of Open Source Software}
}

@article{erickson2025tabarena,
  title={{TabArena}: A living benchmark for machine learning on tabular data},
  author={Erickson, Nick and Purucker, Lennart and Tschalzev, Andrej and Holzm{\"u}ller, David and Desai, Prateek Mutalik and Salinas, David and Hutter, Frank},
  journal={arXiv preprint arXiv:2506.16791},
  year={2025}
}

@article{izbicki2026benchmarking,
  title={Benchmarking Tabular Foundation Models for Conditional Density Estimation in Regression},
  author={Izbicki, Rafael and Rodrigues, Pedro LC},
  journal={arXiv preprint arXiv:2603.26611},
  year={2026}
}

@article{liu2025talent,
  title={TALENT: A tabular analytics and learning toolbox},
  author={Liu, Si-Yang and Cai, Hao-Run and Zhou, Qi-Le and Yin, Huai-Hong and Zhou, Tao and Jiang, Jun-Peng and Ye, Han-Jia},
  journal={Journal of Machine Learning Research},
  volume={26},
  number={226},
  pages={1--16},
  year={2025}
}

@article{Gneiting_sharpness_calibration2007,
    author = {Gneiting, Tilmann and Balabdaoui, Fadoua and Raftery, Adrian E.},
    title = {Probabilistic Forecasts, Calibration and Sharpness},
    journal = {Journal of the Royal Statistical Society Series B: Statistical Methodology},
    volume = {69},
    number = {2},
    pages = {243-268},
    year = {2007},
    month = {03},
    issn = {1369-7412},
    doi = {10.1111/j.1467-9868.2007.00587.x},
    url = {https://doi.org/10.1111/j.1467-9868.2007.00587.x},
    eprint = {https://academic.oup.com/jrsssb/article-pdf/69/2/243/49794500/jrsssb_69_2_243.pdf},
}

@article{hollmann2025accurate,
  title={Accurate predictions on small data with a tabular foundation model},
  author={Hollmann, Noah and M{\"u}ller, Samuel and Purucker, Lennart and Krishnakumar, Arjun and K{\"o}rfer, Max and Hoo, Shi Bin and Schirrmeister, Robin Tibor and Hutter, Frank},
  journal={Nature},
  volume={637},
  number={8045},
  pages={319--326},
  year={2025},
  publisher={Nature Publishing Group UK London}
}

@article{hollmann2022tabpfn,
  title={{TabPFN}: A transformer that solves small tabular classification problems in a second},
  author={Hollmann, Noah and M{\"u}ller, Samuel and Eggensperger, Katharina and Hutter, Frank},
  journal={arXiv preprint arXiv:2207.01848},
  year={2022}
}

@article{gneiting2007strictly,
  title={Strictly proper scoring rules, prediction, and estimation},
  author={Gneiting, Tilmann and Raftery, Adrian E},
  journal={Journal of the American statistical Association},
  volume={102},
  number={477},
  pages={359--378},
  year={2007},
  publisher={Taylor \& Francis}
}

@article{gneiting2011comparing,
  title={Comparing density forecasts using threshold-and quantile-weighted scoring rules},
  author={Gneiting, Tilmann and Ranjan, Roopesh},
  journal={Journal of Business \& Economic Statistics},
  volume={29},
  number={3},
  pages={411--422},
  year={2011},
  publisher={Taylor \& Francis}
}

@inproceedings{qu2026tabiclv2,
  title     = {{TabICLv2}: A Better, Faster, Scalable, and Open Tabular Foundation Model},
  author    = {Qu, Jingang and Holzm{\"u}ller, David and Varoquaux, Ga{\"e}l and Le Morvan, Marine},
  booktitle = {International Conference on Machine Learning},
  year      = {2026},
  url       = {https://openreview.net/forum?id=SxsyLjIfWB}
}

@article{gneiting2011making,
  title={Making and evaluating point forecasts},
  author={Gneiting, Tilmann},
  journal={Journal of the American Statistical Association},
  volume={106},
  number={494},
  pages={746--762},
  year={2011},
  publisher={Taylor \& Francis}
}

@article{merkle2013choosing,
  title={Choosing a strictly proper scoring rule},
  author={Merkle, Edgar C.\ and Steyvers, Mark},
  journal={Decision Analysis},
  volume={10},
  number={4},
  pages={292--304},
  year={2013},
  publisher={INFORMS}
}

@article{waghmare2025proper,
  title={Proper scoring rules for estimation and forecast evaluation},
  author={Waghmare, Kartik and Ziegel, Johanna},
  journal={Annual Review of Statistics and Its Application},
  volume={13},
  year={2025},
  publisher={Annual Reviews}
}

@inproceedings{oesterheld2020decision,
  title={Decision Scoring Rules.},
  author={Oesterheld, Caspar and Conitzer, Vincent},
  booktitle={WINE},
  pages={468},
  year={2020}
}

@article{johnstone2011tailored,
  title={Tailored scoring rules for probabilities},
  author={Johnstone, David J and Jose, Victor Richmond R and Winkler, Robert L},
  journal={Decision Analysis},
  volume={8},
  number={4},
  pages={256--268},
  year={2011},
  publisher={INFORMS}
}

@article{juutilainen2012exceedance,
  title={Exceedance probability score: A novel measure for comparing probabilistic predictions},
  author={Juutilainen, I.\ and Tamminen, S and R{\"o}ning, J},
  journal={Journal of Statistical Theory and Practice},
  volume={6},
  number={3},
  pages={452--467},
  year={2012},
  publisher={Springer}
}

@article{buchweitz2026asymmetric,
  title={Asymmetric penalties underlie proper loss functions in probabilistic forecasting},
  author={Buchweitz, Erez and Romano, Jo{\~a}o Vitor and Tibshirani, Ryan J},
  journal={Biometrika},
  pages={asag042},
  year={2026},
  publisher={Oxford University Press}
}

@article{tran2020methods,
  title={Methods for comparing uncertainty quantifications for material property predictions},
  author={Tran, Kevin and Neiswanger, Willie and Yoon, Junwoong and Zhang, Qingyang and Xing, Eric and Ulissi, Zachary W},
  journal={Machine Learning: Science and Technology},
  volume={1},
  number={2},
  pages={025006},
  year={2020},
  publisher={IOP Publishing}
}

@inproceedings{dheur2023large,
  title={A large-scale study of probabilistic calibration in neural network regression},
  author={Dheur, Victor and Taieb, Souhaib Ben},
  booktitle={International Conference on Machine Learning},
  pages={7813--7836},
  year={2023},
  organization={PMLR}
}

@article{demsar2006statistical,
  title={Statistical comparisons of classifiers over multiple data sets},
  author={Dem{\v{s}}ar, Janez},
  journal={Journal of Machine learning research},
  volume={7},
  number={Jan},
  pages={1--30},
  year={2006}
}

@article{lazic2010problem,
  title={The problem of pseudoreplication in neuroscientific studies: is it affecting your analysis?},
  author={Lazic, Stanley E},
  journal={BMC neuroscience},
  volume={11},
  number={1},
  pages={5},
  year={2010},
  publisher={Springer}
}

@inproceedings{romano2006appropriate,
  title={Appropriate statistics for ordinal level data: Should we really be using t-test and {C}ohen’sd for evaluating group differences on the {NSSE} and other surveys},
  author={Romano, Jeanine and Kromrey, Jeffrey D and Coraggio, Jesse and Skowronek, Jeff},
  booktitle={annual meeting of the Florida Association of Institutional Research},
  volume={177},
  number={34},
  year={2006}
}

@book{nemenyi1963distribution,
  title={Distribution-free multiple comparisons.},
  author={Nemenyi, Peter Bjorn},
  year={1963},
  publisher={Princeton University}
}

@article{jiang2026omnitabbench,
  title={{OmniTabBench}: Mapping the Empirical Frontiers of {GBDT}s, Neural Networks, and Foundation Models for Tabular Data at Scale},
  author={Jiang, Dihong and Cao, Ruoqi and Dang, Zhiyuan and Huang, Li and Zhang, Qingsong and Wang, Zhiyu and Piao, Shihao and Zhu, Shenggao and Chang, Jianlong and Lin, Zhouchen and others},
  journal={arXiv preprint arXiv:2604.06814},
  year={2026}
}

@article{gijsbers2024amlb,
  title={Amlb: an automl benchmark},
  author={Gijsbers, Pieter and Bueno, Marcos LP and Coors, Stefan and LeDell, Erin and Poirier, S{\'e}bastien and Thomas, Janek and Bischl, Bernd and Vanschoren, Joaquin},
  journal={Journal of Machine Learning Research},
  volume={25},
  number={101},
  pages={1--65},
  year={2024}
}

@article{grinsztajn2022tree,
  title={Why do tree-based models still outperform deep learning on typical tabular data?},
  author={Grinsztajn, L{\'e}o and Oyallon, Edouard and Varoquaux, Ga{\"e}l},
  journal={Advances in neural information processing systems},
  volume={35},
  pages={507--520},
  year={2022}
}

@article{olson2017pmlb,
  title={PMLB: a large benchmark suite for machine learning evaluation and comparison},
  author={Olson, Randal S and La Cava, William and Orzechowski, Patryk and Urbanowicz, Ryan J and Moore, Jason H},
  journal={BioData mining},
  volume={10},
  number={1},
  pages={36},
  year={2017},
  publisher={Springer}
}

@article{derrac2015keel,
  title={Keel data-mining software tool: Data set repository, integration of algorithms and experimental analysis framework},
  author={Derrac, J and Garcia, S and Sanchez, L and Herrera, F},
  journal={J. Mult. Valued Logic Soft Comput},
  volume={17},
  pages={255--287},
  year={2015}
}

@inproceedings{fischer2023openml,
  title={Open{ML}-{CTR23}--a curated tabular regression benchmarking suite},
  author={Fischer, Sebastian Felix and Feurer, Matthias and Bischl, Bernd},
  booktitle={AutoML Conference 2023 (Workshop)},
  year={2023}
}

@article{vanschoren2014openml,
  title={{OpenML}: networked science in machine learning},
  author={Vanschoren, Joaquin and Van Rijn, Jan N and Bischl, Bernd and Torgo, Luis},
  journal={ACM SIGKDD Explorations Newsletter},
  volume={15},
  number={2},
  pages={49--60},
  year={2014},
  publisher={ACM New York, NY, USA}
}

@article{bischl2025openml,
  title={{OpenML}: Insights from 10 years and more than a thousand papers},
  author={Bischl, Bernd and Casalicchio, Giuseppe and Das, Taniya and Feurer, Matthias and Fischer, Sebastian and Gijsbers, Pieter and Mukherjee, Subhaditya and M{\"u}ller, Andreas C and N{\'e}meth, L{\'a}szl{\'o} and Oala, Luis and others},
  journal={Patterns},
  volume={6},
  number={7},
  year={2025},
  publisher={Elsevier}
}

@article{rudemo1982empirical,
  title={Empirical choice of histograms and kernel density estimators},
  author={Rudemo, Mats},
  journal={Scandinavian Journal of Statistics},
  pages={65--78},
  year={1982},
  publisher={JSTOR}
}

@article{ghosh2013robust,
  title={Robust estimation for independent non-homogeneous observations using density power divergence with applications to linear regression},
  author={Ghosh, Abhik and Basu, Ayanendranath},
  year={2013}
}

@inproceedings{bostrom2024,
  title={Conformal Prediction in Python with crepes},
  author={Bostr{\"o}m, Henrik},
  booktitle={Proc. of the 13th Symposium on Conformal and Probabilistic Prediction with Applications},
  pages={236--249},
  year={2024},
  organization={PMLR}
}

@article{eo2026exaone,
  title={{EXAONE} {T}abular 1.0: {T}echnical {R}eport},
  author={Eo, Moonjung and Suh, Min-Kook and Cho, Hye-Seung and Kim, Jiwon and Kim, Seoyoon and Nam, Sangjun and Lee, Soonyoung},
  journal={arXiv preprint arXiv:2608.25774},
  year={2026}
}

@article{hosseinzadeh2026tabdpt,
  title={{TabDPT}-{T}urbo: {E}fficient {I}n-{C}ontext {L}earning for {T}abular {P}rediction},
  author={Hosseinzadeh, Rasa and Labach, Alex and Xue, Zexin and Han, Shuyi and Thomas, Valentin and Caterini, Anthony L},
  journal={arXiv preprint arXiv:2608.01400},
  year={2026}
}

@article{tao2026mitra,
  title={Mitra-v2 Technical Report},
  author={Tao, Yefan and Zhang, Xiyuan and Liu, Xinyi and Han, Boran and Maddix, Danielle and Fang, Haoyang and Han, Zhen and Gai, Jiading and Liu, Xuanqing and Bohlke-Schneider, Michael and others},
  journal={arXiv preprint arXiv:2609.04540},
  year={2026}
}

@article{zhang2026limix,
  title={LimiX-2: A Contextual Mechanism Network Towards General Structured-Data Intelligence},
  author={Zhang, Xingxuan and Ren, Gang and Yuan, Hao and Zou, Hao and Tan, Hongze and Wang, Hui and Song, Jianhao and Li, Jiansheng and Zhang, Jiayao and Zhang, Jinghan and others},
  journal={arXiv preprint arXiv:2609.17488},
  year={2026}
}

@article{cho2026causilo,
      title={Causilo Technical Report}, 
      author={Minyong Cho and Minho Jeong and Dooho Lee and Jinmo Lee and Jaemin Yoo},
      year={2026},
      eprint={2609.22866},
      archivePrefix={arXiv},
      primaryClass={cs.LG},
      url={https://arxiv.org/abs/2609.22866}, 
}

@article{wang2026tabldm,
  title={{Xiaomi-TabLDM}: {A} {T}abular {F}oundation {M}odel {T}echnical {R}eport},
  author={Team, TabLDM and Wang, Penghui and Liu, Wei and Wang, Hong and Huang, Chengyue and Sun, Yuxi and Wang, Zirui and Huang, Hongming and Wang, Quan and Liu, Chunxiao and others},
  journal={arXiv preprint arXiv:2609.03880},
  year={2026}
}

@article{gneiting2021mcmc,
author = {Krüger, Fabian and Lerch, Sebastian and Thorarinsdottir, Thordis and Gneiting, Tilmann},
title = {Predictive Inference Based on Markov Chain Monte Carlo Output},
journal = {International Statistical Review},
volume = {89},
number = {2},
pages = {274-301},
doi = {https://doi.org/10.1111/insr.12405},
url = {https://onlinelibrary.wiley.com/doi/abs/10.1111/insr.12405},
eprint = {https://onlinelibrary.wiley.com/doi/pdf/10.1111/insr.12405},
year = {2021}
}

@software{synthefy_2026_20710462,
  author       = {Synthefy and
                  Li, Po-han and
                  Narayanan, Aditya and
                  Narasimhan, Sai Shankar and
                  Mallampalli, Raghav and
                  Agrawal, Aahan and
                  Ajan, Bekzat and
                  Shah, Raimi and
                  Agarwal, Shubhankar},
  title        = {Synthefy Nori: Tabular Foundation Model for Regression},
  month        = jun,
  year         = 2026,
  publisher    = {Zenodo},
  version      = {0.6.0},
  doi          = {10.5281/zenodo.20710462},
  url          = {https://doi.org/10.5281/zenodo.20710462},
}

@article{grinsztajn2026tabpfn,
  title={{TabPFN-3}: {T}echnical {R}eport},
  author={Grinsztajn, L{\'e}o and Fl{\"o}ge, Klemens and Key, Oscar and Birkel, Felix and Jund, Philipp and Roof, Brendan and Manium, Mihir and Hoo, Shi Bin and B{\"u}hler, Magnus and Garg, Anurag and others},
  journal={arXiv preprint arXiv:2605.13986},
  year={2026}
}

@article{shao2024language,
  title={Language generation with strictly proper scoring rules},
  author={Shao, Chenze and Meng, Fandong and Liu, Yijin and Zhou, Jie},
  journal={arXiv preprint arXiv:2405.18906},
  year={2024}
}

@article{scheuerer2015variogram,
  title={Variogram-based proper scoring rules for probabilistic forecasts of multivariate quantities},
  author={Scheuerer, Michael and Hamill, Thomas M},
  journal={Monthly Weather Review},
  volume={143},
  number={4},
  pages={1321--1334},
  year={2015}
}

@article{pinson2013discrimination,
  title={Discrimination ability of the energy score},
  author={Pinson, Pierre and Tastu, Julija},
  year={2013},
  publisher={Technical University of Denmark}
}
